\documentclass[10pt,journal,letterpaper,compsoc]{IEEEtran}

\usepackage{graphicx}
\usepackage{amsmath,amssymb} % define this before the line numbering.
\usepackage{color}
\usepackage{epsfig}
\usepackage{subfigure}
\usepackage{verbatim}
\usepackage{booktabs}
\usepackage{multirow}
\usepackage{float}
\usepackage{wrapfig}

\usepackage{times}
\usepackage{bm}
\usepackage{array}
\usepackage{makecell}
\usepackage{stackengine}
\usepackage[font={small}]{caption}

\usepackage[pagebackref=true,letterpaper=true,colorlinks,bookmarks=false]{hyperref}
\makeatletter

\newcommand{\Rmnum}[1]{\expandafter\@slowromancap\romannumeral #1@}
\makeatother

\begin{document}

\title{StackGAN++: Realistic Image Synthesis \\with Stacked Generative Adversarial Networks}

\author{Han Zhang,
        \;Tao Xu,
        \;Hongsheng Li,
        \;Shaoting Zhang,~\IEEEmembership{Senior Member,~IEEE},
        \\Xiaogang  Wang,~\IEEEmembership{Member,~IEEE},
        \;Xiaolei Huang,~\IEEEmembership{Member,~IEEE},
        \;Dimitris N.~Metaxas$^{\ast}$,~\IEEEmembership{Fellow,~IEEE}% <-this % stops a space
\IEEEcompsocitemizethanks{
\IEEEcompsocthanksitem H. Zhang is with the Department of Computer Science, Rutgers University, Piscataway, NJ, 08854. E-mail: han.zhang@cs.rutgers.edu\protect\\%
\vspace{-8pt}
\IEEEcompsocthanksitem T. Xu is with the Department of Computer Science and Engineering, Lehigh University, Bethlehem, PA, 18015. E-mail: tax313@lehigh.edu\protect\\%
\vspace{-8pt}
\IEEEcompsocthanksitem H. Li is with the Department of Electronic Engineering, The Chinese University of Hong Kong, Shatin, N.~T., Hong Kong. E-mail: hsli@ee.cuhk.edu.hk\protect\\%
\vspace{-8pt}
\IEEEcompsocthanksitem S. Zhang is with the Department of Computer Science, University of North Carolina at Charlotte, Charlotte, NC 28223. E-mail: szhang16@uncc.edu\protect\\%
\vspace{-8pt}
\IEEEcompsocthanksitem X. Wang is with the Department of Electronic Engineering, The Chinese University of Hong Kong, Shatin, N.~T., Hong Kong. E-mail: xgwang@ee.cuhk.edu.hk\protect\\%
\vspace{-8pt}
\IEEEcompsocthanksitem X. Huang is with the Department of Computer Science and Engineering, Lehigh University, Bethlehem, PA, 18015. E-mail: xih206@lehigh.edu\protect\\%
\vspace{-8pt}
\IEEEcompsocthanksitem D.~N. Metaxas is with the Department of Computer Science, Rutgers University, Piscataway, NJ, 08854. E-mail: dnm@cs.rutgers.edu}% <-this % stops an unwanted space
\thanks{
The first two authors contributed equally to this work. Asterisk
indicates the corresponding author.
% Manuscript received April 19, 2005; revised August 26, 2015.
}
}

\IEEEcompsoctitleabstractindextext{%
\begin{abstract}
Although Generative Adversarial Networks (GANs) have shown remarkable success in various tasks, they still face challenges in generating high quality images. In this paper, we propose Stacked Generative Adversarial Networks (StackGANs) aimed at generating high-resolution photo-realistic images.   First, we propose a two-stage generative adversarial network architecture, StackGAN-v1, for text-to-image synthesis. The Stage-\Rmnum{1} GAN sketches the primitive shape and colors of a scene based on a given text description, yielding low-resolution images. The Stage-\Rmnum{2} GAN takes Stage-\Rmnum{1} results and the text description as inputs, and generates high-resolution images with photo-realistic details. Second, an advanced multi-stage generative adversarial network architecture, StackGAN-v2, is proposed for both conditional and unconditional generative tasks. Our StackGAN-v2 consists of multiple generators and multiple discriminators arranged in a tree-like structure; images at multiple scales corresponding to the same scene are generated from different branches of the tree. StackGAN-v2 shows more stable training behavior than StackGAN-v1 by jointly approximating multiple distributions. Extensive experiments demonstrate that the proposed stacked generative adversarial networks significantly outperform other state-of-the-art methods in generating photo-realistic images. 
\end{abstract}

}

\maketitle

\IEEEdisplaynotcompsoctitleabstractindextext

\IEEEpeerreviewmaketitle

%%%%%%%%% BODY TEXT
\hyphenpenalty=1000

%-------------------------------------------------------------------------
\section{Introduction}
{
Generative Adversarial Networks (GANs) were proposed by Goodfellow \emph{et al}.~\cite{goodfellow2014generative}. In the original setting, GANs are composed of a generator and a discriminator that are trained with competing goals. The generator is trained to generate samples towards the true data distribution to fool the discriminator, while the discriminator is optimized to distinguish between real samples from the true data distribution and fake samples produced by the generator. Recently, GANs have shown great potential in simulating complex data distributions, such as those of texts~\cite{che2017}, images~\cite{Radford15} and videos~\cite{VondrickPT16}. 
}

{
Despite the success, GANs are known to be difficult to train. The training process is usually unstable and sensitive to the choices of hyper-parameters. Several papers argued that the instability is partially due to the disjoint supports of the data distribution and the implied model distribution~\cite{Casper2016, ArjovskyB17}. This problem is more severe when training GANs to generate high-resolution (\emph{e.g}., 256$\times$256) images because the chance is very low for the image distribution and the model distribution to share supports in a high-dimensional space. Moreover, a common failure phenomenon for GANs training is \emph{mode collapse}, where many of the generated samples contain the same color or texture pattern. 
}

{
In order to stabilize the GANs' training process and improve sample diversity, several methods tried to address the challenges by proposing new network architectures~\cite{Radford15}, introducing heuristic tricks~\cite{Salimans2016} or modifying the learning objectives~\cite{Martin17WGAN,CheLJBL16,Salimans18}. But most of the previous methods are designed to approximate the image distribution at a single scale. Due to the difficulty in directly approximating the high-resolution image data distribution, most previous methods are limited to generating low-resolution images. To circumvent this difficulty, we observe that, real world data, especially natural images, can be modeled at different scales \cite{ruderman1994statistics}. One can view multi-resolution digitized images as samples from the same continuous image signal with different sampling rates. Henceforth, the distributions of images at multiple discrete scales are related. Apart from multiple distributions of different scales, images coupled with or without auxiliary conditioning variables (\emph{e.g}., class labels or text descriptions) can be viewed as conditional distributions or unconditional distributions, which are also related distributions. Motivated by these observations, we argue that GANs can be stably trained to generate high resolution images by breaking the difficult generative task into sub-problems with progressive goals. Thus, we propose Stacked Generative Adversarial Networks (StackGANs) to model a series of low-to-high-dimensional data distributions. 
}

{
First,  we  propose  a two-stage generative adversarial network, StackGAN-v1, to generate images from text descriptions through a sketch-refinement process~\cite{Han16}. Low-resolution images are first generated by our Stage-\Rmnum{1} GAN. On top of the Stage-\Rmnum{1} GAN, we stack Stage-\Rmnum{2} GAN to generate high-resolution (\emph{e.g}., 256$\times$256) images. By conditioning on the Stage-\Rmnum{1} result and the text again, Stage-\Rmnum{2} GAN learns to capture the text information that is omitted by Stage-\Rmnum{1} GAN and draws more details. Further, we propose a novel Conditioning Augmentation (CA) technique to encourage smoothness in the latent conditioning manifold~\cite{Han16}. It allows small random perturbations in the conditioning manifold and increases the diversity of synthesized images. 
}

{
Second, we propose an advanced multi-stage generative adversarial network architecture, StackGAN-v2, for both conditional and unconditional generative tasks. StackGAN-v2 has multiple generators that share most of their parameters in a tree-like structure. As shown in Fig.~\ref{fig:StackGAN-v2}, the input to the network can be viewed as the root of the tree, and multi-scale images are generated from different branches of the tree. The generator at the deepest branch has the final goal of generating photo-realistic high-resolution images. Generators at intermediate branches have progressive goals of generating small to large images to help accomplish the final goal. The whole network is jointly trained to approximate different but highly related image distributions at different branches. The positive feedback from modeling one distribution can improve the learning of others. For conditional image generation tasks, our proposed StackGAN-v2 simultaneously approximates the unconditional image-only distribution and the image distribution conditioned on text descriptions. Those two types of distributions are complementary to each other. Moreover, we propose a color-consistency regularization term to guide our generators to generate more coherent samples across different scales. The regularization provides additional constraints to facilitate multi-distribution approximation, which is especially useful in the unconditional setting where there is no instance-wise supervision between the image and the input noise vector. 
}

{
In summary, the proposed Stacked Generative Adversarial Networks have three major contributions. (\textit{i}) Our StackGAN-v1 for the first time generates images of 256$\times$256 resolution with photo-realistic details from text descriptions. (\textit{ii}) A new Conditioning Augmentation technique is proposed to stabilize the conditional GANs' training and also improve the diversity of the generated samples. (\textit{iii}) Our StackGAN-v2 further improves the quality of generated images and stabilizes the GANs' training by jointly approximating multiple distributions. In the remainder of this paper, we first discuss related work and preliminaries in section~\ref{sec:related} and section~\ref{sec:bg}, respectively. We then introduce our StackGAN-v1~\cite{Han16} in section~\ref{sec:v1} and StackGAN-v2 in section~\ref{sec:v2}. In section~\ref{sec:exp}, extensive experiments are conducted to evaluate the proposed methods. Finally, we make conclusions in section~\ref{sec:conclusion}. The source code for StackGAN-v1 is available at {\href{https://github.com/hanzhanggit/StackGAN}{https://github.com/hanzhanggit/StackGAN}}, and the source code for StackGAN-v2 is available at {\href{https://github.com/hanzhanggit/StackGAN-v2}{https://github.com/hanzhanggit/StackGAN-v2}}.
}
%-------------------------------------------------------------------------

\section{Related Work}\label{sec:related}
{
Generative image modeling is a fundamental problem in computer vision. There has been remarkable progress in this direction with the emergence of deep learning techniques. Variational Autoencoders (VAEs)~\cite{KingmaW14, RezendeMW14} formulate the problem with probabilistic graphical models with the goal of maximizing the lower bound of data likelihood. Autoregressive models (\emph{e.g}., PixelRNN)~\cite{OordKK16} that utilize neural networks to model the conditional distribution of the pixel space have also generated appealing synthetic images. Recently, Generative Adversarial Networks (GANs)~\cite{goodfellow2014generative} have shown promising performance for generating sharper images. But the training instability makes it hard for GANs to generate high-resolution (\emph{e.g}., 256$\times$256) images. A lot of works have been proposed to stabilize the training and improve the image qualities~\cite{Radford15, Salimans2016, MetzICLR17, Zhao2016, CheLJBL16, NguyenYBDC17}. 
}

{
Built upon these generative models, conditional image generation has also been studied. Most methods utilize simple conditioning variables such as attributes or class labels~\cite{YanYSL16, Oord16, ChenDHSSA16, Odena2016}. There are also works conditioned on images to generate images, including photo editing~\cite{Brock2016, ZhuKSE16}, domain transfer~\cite{Taigmaniclr17, pix2pix2017} and super-resolution~\cite{Casper2016, Christian2016}. However, super-resolution methods~\cite{Casper2016, Christian2016} can only add limited details to low-resolution images and can not correct large defects. In contrast, the latter stages in our proposed StackGANs can not only add details to low-resolution images generated by earlier stages but also correct defects in them. Recently, several methods have been developed to generate images from unstructured text. Mansimov \emph{et al}.~\cite{MansimovPBS15} built an AlignDRAW model by learning to estimate alignment between text and the generating canvas.  Reed \emph{et al}.~\cite{reed2016iclr17} used conditional PixelCNN to generate images using text descriptions and object location constraints. Nguyen \emph{et al}.~\cite{NguyenYBDC17} used an approximate Langevin sampling approach to generate images conditioned on text. However, their sampling approach requires an inefficient iterative optimization process. With conditional GANs, Reed \emph{et al}.~\cite{reed2016generative} successfully generated plausible 64$\times$64 images for birds and flowers based on text descriptions. Their follow-up work~\cite{reed2016learning} was able to generate 128$\times$128 images by utilizing additional annotations on object part locations.
}

{
Given the difficulties in modeling details of natural images, many works have been proposed to use multiple GANs to improve sample quality. Wang \emph{et al}.~\cite{WangG16} utilized a structure GAN and a style GAN to synthesize images of indoor scenes. Yang \emph{et al}.~\cite{YangKBP17} factorized image generation into foreground and background generation with layered recursive GANs. Huang \emph{et al}.~\cite{huang2016sgan} added several GANs to reconstruct the multi-level representations of a pre-trained discriminative model. But they were unable to generate high resolution images with photo-realistic details. Durugkar \emph{et al}.~\cite{DurugkarGM17} used multiple discriminators along with one generator to increase the chance of the generator receiving effective feedback. However, all discriminators in their framework are trained to approximate the image distribution at a single scale. Some methods~\cite{DentonCSF15,KarrasALL18} follow the same intuition as our work. We all agree that it is beneficial to break the high-resolution image generation task into several easier subtasks to be accomplished in multiple stages. Denton \emph{et al}.~\cite{DentonCSF15} built a series of GANs within a Laplacian pyramid framework (LAPGANs). At each level of the pyramid, a residual image conditioned on the image of the previous stage is generated and then added back to the input image to produce the input for the next stage. Instead of producing a residual image, our StackGANs directly generate high resolution images that are conditioned on their low-resolution inputs. Concurrent to our work, Kerras \emph{et al}.~\cite{KarrasALL18} incrementally add more layers in the generator and discriminator for high resolution image generation. The main difference in terms of experimental setting is that they used a more restrained upsampling rule: starting from 4$\times$4 pixels, their image resolution is increased by a factor of 2 between consecutive image generation stages.  Furthermore, although StackGANs, LAPGANs~\cite{DentonCSF15} and Progressive GANs~\cite{KarrasALL18} all put emphasis on adding finer details in higher resolution images, our StackGANs can also correct incoherent artifacts or defects in low resolution results by utilizing an encoder-decoder network before the upsampling layers. 
}
%-------------------------------------------------------------------------

\section{Preliminaries}\label{sec:bg}
{
Generative Adversarial Networks (GANs)~\cite{goodfellow2014generative} are composed of two models that are alternatively trained to compete with each other. The generator $G$ is optimized to reproduce the true data distribution $p_{data}$ by generating images that are difficult for the discriminator $D$ to differentiate from real images. Meanwhile, $D$ is optimized to distinguish real images and synthetic images generated by $G$. Overall, the training procedure is a minmax two-player game with the following objective function, 
\begin{equation}\label{eq:GAN_ori}
\begin{aligned}
\min_{G} \max_{D} V(D,G) = \; & \mathbb{E}_{x \sim {p_{data}}} [\log D(x)] \; + \\
& \mathbb{E}_{z \sim {p_{z}}} [\log(1 - D(G(z)))],
\end{aligned}
\end{equation}
where $x$ is a real image from the true data distribution $p_{data}$, and $z$ is a noise vector sampled from the prior distribution $p_{z}$ (\emph{e.g}., uniform or Gaussian distribution). In practice, the generator $G$ is modified to maximize $\log(D(G(z)))$ instead of minimizing $\log(1-D(G(z)))$ to mitigate the problem of gradient vanishing~\cite{goodfellow2014generative}. We use this modified non-saturating objective in all our experiments.

Conditional GANs~\cite{gauthier2015conditional, Mirza14} are extension of GANs where both the generator and discriminator receive additional conditioning variables $c$, yielding $G(z,c)$ and $D(x,c)$. This formulation allows $G$ to generate images conditioned on variables $c$. 
}
%-------------------------------------------------------------------------

%
%
\begin{figure*}[tb]
\begin{center}
\includegraphics[width=0.85\linewidth]{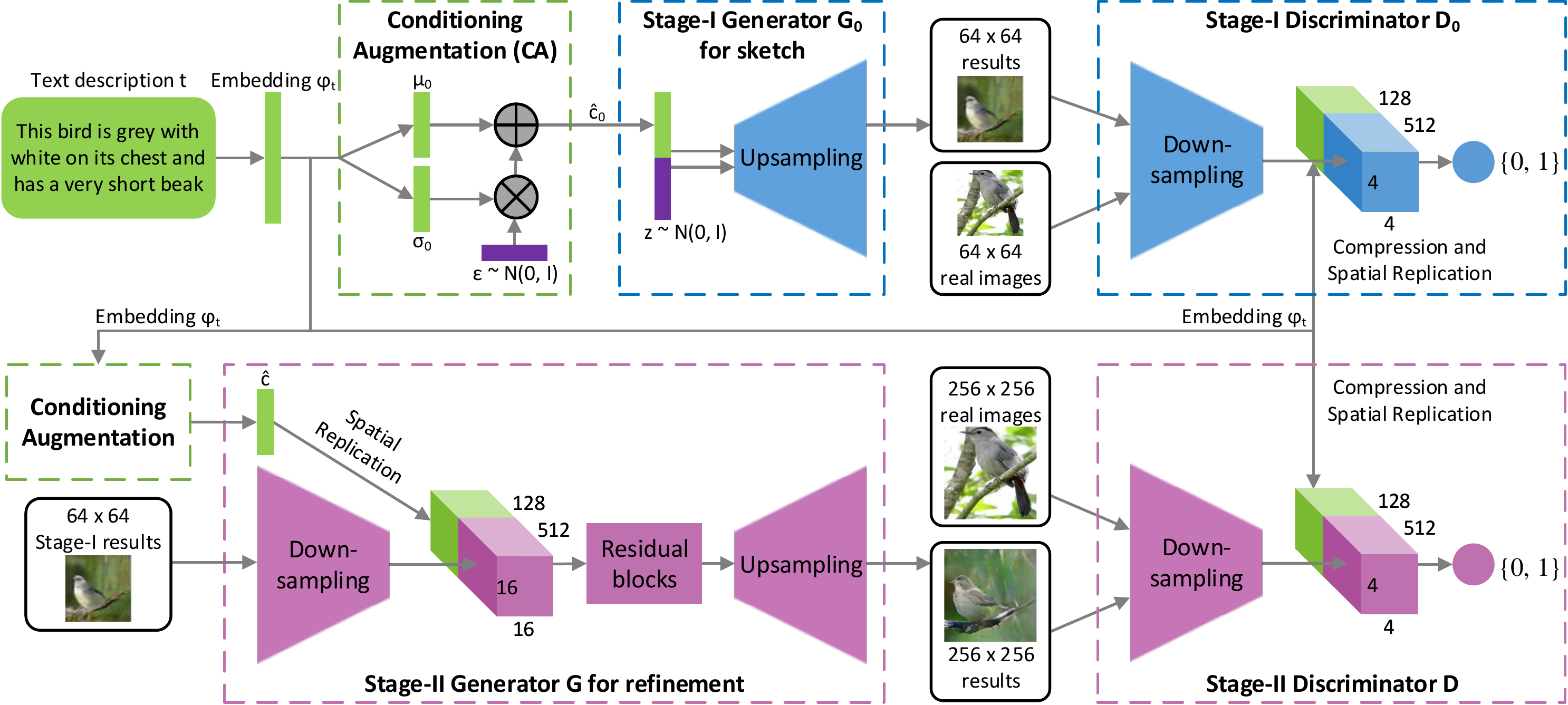}
\end{center}
 \vspace{-5pt}
   \caption{The architecture of the proposed StackGAN-v1. The Stage-\Rmnum{1} generator draws a low-resolution image by sketching rough shape and basic colors of the object from the given text and painting the background from a random noise vector. Conditioned on Stage-\Rmnum{1} results, the Stage-\Rmnum{2} generator corrects defects and adds compelling details into Stage-\Rmnum{1} results, yielding a more realistic high-resolution image.
   }
	\vspace{-10pt}
\label{fig:StackGAN-v1}
\end{figure*}

\section{StackGAN-v1: Two-stage Generative Adversarial Network}\label{sec:v1}

To generate high-resolution images with photo-realistic details, we propose a simple yet effective two-stage generative adversarial network, StackGAN-v1. As shown in Fig.~\ref{fig:StackGAN-v1}, it decomposes the text-to-image generative process into two stages.   Stage-\Rmnum{1} GAN sketches the primitive shape and basic colors of the object conditioned on the given text description, and draws the background layout from a random noise vector, yielding a low-resolution image.   Stage-\Rmnum{2} GAN corrects defects in the low-resolution image from Stage-I and completes details of the object by reading the text description again, producing a high-resolution photo-realistic image.

\subsection{Conditioning Augmentation}
As shown in Fig.~\ref{fig:StackGAN-v1}, the text description $t$ is first encoded by an encoder, yielding a text embedding $\varphi_{t}$. In previous works~\cite{reed2016generative,reed2016learning}, the text embedding is nonlinearly transformed to generate conditioning latent variables as the input of the generator. However, latent space for the text embedding is usually high dimensional ($>100$ dimensions). With limited amount of data, it usually causes discontinuity in the latent data manifold, which is not desirable for learning the generator. To mitigate this problem, we introduce a Conditioning Augmentation technique to produce additional conditioning variables $\hat{c}$. In contrast to the fixed conditioning text variable $c$ in \cite{reed2016generative,reed2016learning}, we randomly sample the latent variables $\hat{c}$ from an independent Gaussian distribution $\mathcal{N}(\mu(\varphi_{t}), \Sigma(\varphi_{t}))$, where the mean $\mu(\varphi_{t})$ and diagonal covariance matrix $\Sigma(\varphi_{t})$ are functions of the text embedding $\varphi_{t}$. The proposed Conditioning Augmentation yields more training pairs given a small number of image-text pairs, and thus encourages robustness to small perturbations along the conditioning manifold. To further enforce the smoothness over the conditioning manifold and avoid overfitting~\cite{Doersch16, LarsenSLW16}, we add the following regularization term to the objective of the generator during training, 
\begin{equation}\label{eq:GaussianKL}
\begin{aligned}
D_{KL}\left(\mathcal{N}(\mu(\varphi_{t}), \Sigma(\varphi_{t})) \, || \, \mathcal{N}(0, I)\right),
\end{aligned}
\end{equation}
which is the Kullback-Leibler divergence (KL divergence) between the standard Gaussian distribution and the conditioning Gaussian distribution. The randomness introduced in the Conditioning Augmentation is beneficial for modeling text to image translation as the same sentence usually corresponds to objects with various poses and appearances.

\subsection{Stage-\Rmnum{1} GAN}
Instead of directly generating a high-resolution image conditioned on the text description, we simplify the task to first generate a low-resolution image with our Stage-\Rmnum{1} GAN, which focuses on drawing only rough shape and correct colors for the object.

Let $\varphi_{t}$ be the text embedding of the given description. The Gaussian conditioning variables $\hat{c}_0$ for text embedding are sampled from $\mathcal{N}(\mu_0(\varphi_{t}), \Sigma_0(\varphi_{t}))$ to capture the meaning of $\varphi_{t}$ with variations. Conditioned on $\hat{c}_0$ and random variable $z$, Stage-\Rmnum{1} GAN trains the discriminator $D_0$ and the generator $G_0$ by alternatively maximizing $\mathcal{L}_{D_0}$ in Eq.~(\ref{eq:D_0}) and minimizing $\mathcal{L}_{G_0}$ in Eq.~(\ref{eq:G_0}), 
\begin{equation}\label{eq:D_0}
\begin{aligned}
 \mathcal{L}_{D_0} =  &\; \mathbb{E}_{(I_{0},t) \sim {p_{data}}} [\log D_{0}(I_{0}, \varphi_{t})] \; + \\
&\; \mathbb{E}_{z \sim {p_{z}}, t \sim p_{data}} [\log(1 - D_{0}(G_{0}(z,\hat{c}_0), \varphi_{t}))],
\end{aligned}
\end{equation}
\begin{equation}\label{eq:G_0}
\begin{aligned}
\mathcal{L}_{G_0} = &\; \mathbb{E}_{z \sim {p_{z}}, t \sim p_{data}} [-\log D_{0}(G_{0}(z,\hat{c}_0), \varphi_{t})] \; + \\
                    &\; \lambda D_{KL}(\mathcal{N}(\mu_0(\varphi_{t}), \Sigma_0(\varphi_{t})) \, || \, \mathcal{N}(0, I)), 
\end{aligned}
\end{equation}
where the real image $I_0$ and the text description $t$ are from the true data distribution $p_{data}$. $z$ is a noise vector randomly sampled from a given distribution $p_z$ (Gaussian distribution in this paper). $\lambda$ is a regularization parameter that balances the two terms in Eq.~(\ref{eq:G_0}). We set $\lambda=1$ for all our experiments.  Using the reparameterization trick introduced in~\cite{KingmaW14}, both $\mu_0(\varphi_{t})$ and $\Sigma_0(\varphi_{t})$ are learned jointly with the rest of the network.  To extract a visually-discriminative text embedding of the given description, we follow the approach of Reed \emph{et al}.~\cite{reed2016cvpr} to pre-train a text encoder. It is a character level CNN-RNN model that maps text descriptions to the common feature space of images by learning a correspondence function between texts with images~\cite{reed2016cvpr}.

\textbf{Model Architecture. }
For the generator $G_0$, to obtain text conditioning variable $\hat{c}_0$, the text embedding $\varphi_{t}$ is first fed into a fully connected layer to generate $\mu_0$ and $\sigma_0$ ($\sigma_0$ are the values in the diagonal of $\Sigma_0$) for the Gaussian distribution $\mathcal{N}(\mu_0(\varphi_{t}), \Sigma_0(\varphi_{t}))$. $\hat{c}_0$ are then sampled from the Gaussian distribution.
Our $N_g$ dimensional conditioning vector $\hat{c_0}$ is computed by $\hat{c_0} = \mu_0 + \sigma_0 \odot \epsilon$ (where $\odot$ is the element-wise multiplication, $\epsilon  \sim \mathcal{N}(0, I)$). 
Then, $\hat{c_0}$ is concatenated with a $N_z$ dimensional noise vector to generate a $W_{0} \times H_{0}$ image by a series of up-sampling blocks.  

For the discriminator $D_0$, the text embedding $\varphi_{t}$ is first compressed to $N_d$ dimensions using a fully-connected layer and then spatially replicated to form a $M_d \times M_d \times N_d$ tensor. 
Meanwhile, the image is fed through a series of down-sampling blocks until it has $M_d \times M_d$ spatial dimension. 
Then, the image filter map is concatenated along the channel dimension with the text tensor. 
The resulting tensor is further fed to a 1$\times$1 convolutional layer to jointly learn features across the image and the text. 
Finally, a fully-connected layer with one node is used to produce the decision score.

\subsection{Stage-\Rmnum{2} GAN} 
Low-resolution images generated by Stage-\Rmnum{1} GAN usually lack vivid object parts and might contain shape distortions. Some details in the text might also be omitted in the first stage, which is vital for generating photo-realistic images.  
Our Stage-\Rmnum{2} GAN is built upon Stage-\Rmnum{1} GAN results to generate high-resolution images. 
It is conditioned on low-resolution images and also the text embedding again to correct defects in Stage-\Rmnum{1} results. The Stage-\Rmnum{2} GAN completes previously ignored text information to generate more photo-realistic details.

Conditioning on the low-resolution result $s_0 = G_0(z, \hat{c}_0)$ and Gaussian latent variables $\hat{c}$, the discriminator $D$ and generator $G$ in Stage-\Rmnum{2} GAN are trained by alternatively maximizing $\mathcal{L}_{D}$ in Eq.~(\ref{eq:D}) and minimizing $\mathcal{L}_{G}$ in Eq.~(\ref{eq:G}), 
\begin{equation}\label{eq:D}
\begin{aligned}
\mathcal{L}_{D} =  &\; \mathbb{E}_{(I,t) \sim {p_{data}}} [\log D(I, \varphi_{t})] \; + \\
&\; \mathbb{E}_{s_0 \sim {p_{G_0}}, t \sim p_{data}} [\log(1 - D(G(s_0, \hat{c}), \varphi_{t}))],
\end{aligned}
\end{equation}
\begin{equation}\label{eq:G}
\begin{aligned}
\mathcal{L}_{G} = &\; \mathbb{E}_{s_0 \sim {p_{G_0}}, t \sim p_{data}} \left[-\log D(G(s_0, \hat{c}), \varphi_{t})\right] \; + \\
                    &\; \lambda D_{KL}\left(\mathcal{N}(\mu(\varphi_{t}), \Sigma(\varphi_{t})) \, || \, \mathcal{N}(0, I)\right), 
\end{aligned}
\end{equation}
Different from the original formulation of GANs, the random noise $z$ is not used in this stage with the assumption that the randomness has already been preserved by $s_0$. Gaussian conditioning variables $\hat{c}$ used in this stage and $\hat{c}_0$ used in Stage-\Rmnum{1} GAN share the same pre-trained text encoder, generating the same text embedding $\varphi_{t}$. However, Stage-I and Stage-II Conditioning Augmentation have different fully connected layers for generating different means and standard deviations. In this way, Stage-\Rmnum{2} GAN learns to capture useful information in the text embedding that is omitted by Stage-\Rmnum{1} GAN.

\textbf{Model Architecture. }
We design Stage-\Rmnum{2} generator as an encoder-decoder network with residual blocks~\cite{HeZRS15}. 
Similar to the previous stage, the text embedding $\varphi_{t}$ is used to generate the $N_g$ dimensional text conditioning vector $\hat{c}$, which is spatially replicated to form a $M_{g} \times M_{g} \times N_g$ tensor. 
Meanwhile, the Stage-\Rmnum{1} result $s_0$ generated by Stage-\Rmnum{1} GAN is fed into several down-sampling blocks (\emph{i.e}., encoder) until it has a spatial size of $M_{g} \times M_{g}$. The image features and the text features are concatenated along the channel dimension. 
The encoded image features coupled with text features are fed into several residual blocks, which are designed to learn multi-modal representations across image and text features. 
Finally, a series of up-sampling layers (\emph{i.e}., decoder) are used to generate a $W \times H$ high-resolution image. Such a generator is able to help rectify defects in the input image while add more details to generate the realistic high-resolution image.

For the discriminator, its structure is similar to that of Stage-\Rmnum{1} discriminator with only extra down-sampling blocks since the image size is larger in this stage.  
To explicitly enforce GANs to learn better alignment between the image and the conditioning text, rather than using the vanilla discriminator, we adopt the matching-aware discriminator proposed by Reed \emph{et al}.~\cite{reed2016generative} for both stages. 
During training, the discriminator takes real images and their corresponding text descriptions as positive sample pairs, whereas negative sample pairs consist of two groups. 
The first is real images with mismatched text embeddings, while the second is synthetic images with their corresponding text embeddings. 

\vspace{-2pt}

\subsection{Implementation details}
{
The up-sampling blocks consist of the nearest-neighbor upsampling followed by a 3$\times$3 stride 1 convolution. Batch normalization~\cite{IoffeS15} and ReLU activation are applied after every convolution except the last one.  The residual blocks consist of  3$\times$3 stride 1 convolutions, Batch normalization and ReLU. Two residual blocks are used in 128$\times$128 StackGAN-v1 models while four are used in 256$\times$256 models. The down-sampling blocks consist of 4$\times$4 stride 2 convolutions, Batch normalization and LeakyReLU, except that the first one does not have Batch normalization.

By default, $N_g=128$,  $N_z=100$, $M_g=16$,  $M_d=4$, $N_d=128$,  $W_{0}=H_{0} = 64$ and $W=H=256$. 
For training, we first iteratively train $D_0$ and $G_0$ of Stage-\Rmnum{1} GAN for 600 epochs by fixing Stage-\Rmnum{2} GAN. Then we iteratively train $D$ and $G$ of Stage-\Rmnum{2} GAN for another 600 epochs by fixing Stage-\Rmnum{1} GAN. All networks are trained using ADAM~\cite{KingmaB14} solver with $beta1=0.5$. The batch size is 64, and the learning rate is initialized to be 0.0002 and decayed to $1/2$ of its previous value every 100 epochs. The source code for StackGAN-v1 is available at {\href{https://github.com/hanzhanggit/StackGAN}{https://github.com/hanzhanggit/StackGAN}} for more implementation details.
}
%-------------------------------------------------------------------------

%
%
\begin{figure*}[tb]
\centering
\includegraphics[width=1\linewidth]{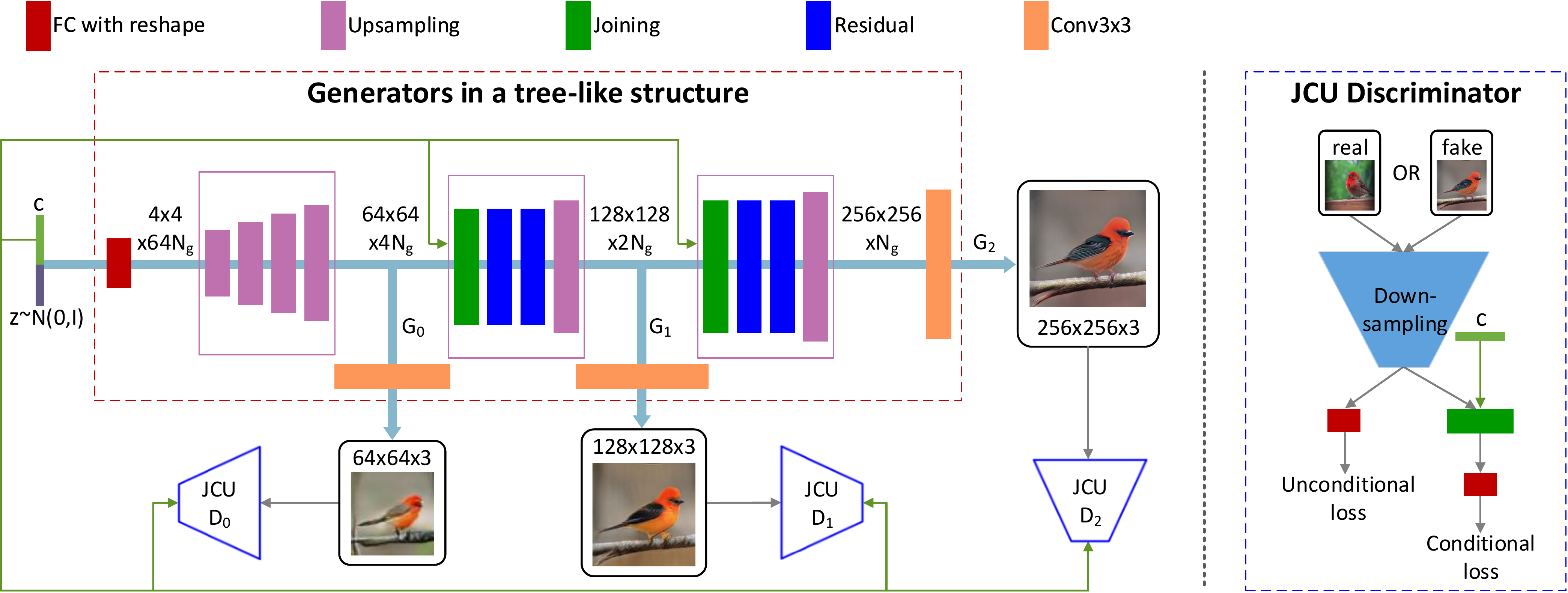}
\vspace{-12pt}
\caption{The overall framework of our proposed StackGAN-v2 for the conditional image synthesis task. $c$ is the vector of conditioning variables which can be computed from the class label, the text description, \emph{etc}.. $N_g$ and $N_d$ are the numbers of channels of a tensor.}
\vspace{-10pt}
\label{fig:StackGAN-v2}
\end{figure*}

\section{StackGAN-v2: Multi-distribution Generative Adversarial Network} \label{sec:v2}

As discussed above, our StackGAN-v1 has two separate networks, Stage-I GAN and Stage-II GAN, to model low-to-high resolution image distributions. To make the framework more general, in this paper, we propose a new end-to-end network, StackGAN-v2, to model a series of multi-scale image distributions. As shown in Fig.~\ref{fig:StackGAN-v2}, StackGAN-v2 consists of multiple generators ($G$s) and discriminators ($D$s) in a tree-like structure. Images from low-resolution to high-resolution are generated from different branches of the tree. At each branch, the generator captures the image distribution at that scale and the discriminator estimates the probability that a sample came from training images of that scale rather than the generator. The generators are jointly trained to approximate the multiple distributions, and the generators and discriminators are trained in an alternating fashion. In this section, we explore two types of multi-distributions: (1) multi-scale image distributions; and (2) joint conditional and unconditional image distributions.

\subsection{Multi-scale image distributions approximation} \label{sec:jcu}

Our StackGAN-v2 framework has a tree-like structure, which takes a noise vector $z \sim p_{noise}$ as the input and has multiple generators to produce images of different scales. The $p_{noise}$ is a prior distribution, which is usually chosen as the standard normal distribution.  The latent variables $z$ are transformed to hidden features layer by layer. We compute the hidden features $h_i$ for each generator $G_i$ by a non-linear transformation,
\begin{equation}\label{eq:hidden1}
h_{0}=F_{0}(z); \quad h_{i} = F_{i}(h_{i-1}, z), \; i = 1, 2, ..., m-1,
\end{equation} 
where $h_{i}$ represents hidden features for the $i^{th}$ branch, $m$ is the total number of branches, and $F_{i}$ are modeled as neural networks (see Fig.~\ref{fig:StackGAN-v2} for illustration). In order to capture information omitted in preceding branches, the noise vector $z$ is concatenated to the hidden features $h_{i-1}$ as the inputs of $F_i$ for calculating $h_{i}$. Based on hidden features at different layers ($h_0$, $h_1$, ..., $h_{m-1}$), generators produce samples of small-to-large scales ($s_0$, $s_1$, ..., $s_{m-1}$),
\begin{equation}\label{eq:generate}
s_i = G_{i}(h_i), \; i = 0, 1, ..., m-1,
\end{equation}
where $G_{i}$ is the generator for the $i^{th}$ branch.

Following each generator $G_{i}$, a discriminator $D_{i}$, which takes a real image $x_i$ or a fake sample $s_i$ as input, is trained to classify inputs into two classes (real or fake) by minimizing the following cross-entropy loss,
\begin{equation}\label{eq:LD}
% \begin{aligned}
\small
\mathcal{L}_{D_{i}} = -\mathbb{E}_{x_{i} \sim {p_{data_i}}} [\log D_{i}(x_{i})] - \mathbb{E}_{s_{i} \sim {p_{G_{i}}}} [ \log(1 - D_{i}(s_{i})],
% \end{aligned}
\end{equation}
where $x_i$ is from the true image distribution $p_{data_i}$ at the $i^{th}$ scale, and $s_i$ is from the model distribution $p_{G_{i}}$ at the same scale. The multiple discriminators are trained in parallel, and each of them focuses on a single image scale.

Guided by the trained discriminators, the generators are optimized to jointly approximate multi-scale image distributions ($p_{data_0}, p_{data_1}, ..., p_{data_{m-1}}$) by minimizing the following loss function,
\begin{equation}\label{eq:LG}
\small
\mathcal{L}_{G} = \sum_{i=1}^m \mathcal{L}_{G_{i}}, \quad \mathcal{L}_{G_{i}} = -\mathbb{E}_{s_{i} \sim {p_{G_{i}}}} \left[\log  D_{i}(s_{i})\right],
\end{equation}
where $\mathcal{L}_{G_{i}}$ is the loss function for approximating the image distribution at the $i^{th}$ scale (\emph{i.e}., $p_{data_i}$). During the training process, the discriminators $D_i$ and the generators $G_i$ are alternately optimized till convergence.  

The motivation of the proposed StackGAN-v2 is that, by modeling data distributions at multiple scales, if any one of those model distributions shares support with the real data distribution at that scale, the overlap could provide good gradient signal to expedite or stabilize training of the whole network at multiple scales. For instance, approximating the low-resolution image distribution at the first branch results in images with basic color and structures. Then the generators at the subsequent branches can focus on completing details for generating higher resolution images.

\subsection{Joint conditional and unconditional distribution approximation}

For unconditional image generation, discriminators in StackGAN-v2 are trained to distinguish real images from fake ones. To handle conditional image generation, conventionally, images and their corresponding conditioning variables are input into the discriminator to determine whether an image-condition pair matches or not, which guides the generator to approximate the conditional image distribution. We propose conditional StackGAN-v2 that jointly approximates conditional and unconditional image distributions.

For the generator of our conditional StackGAN-v2, $F_0$ and $F_i$ are converted to take the conditioning vector $c$ as input, such that $h_{0}=F_{0}(c,z)$ and $h_{i}=F_{i}(h_{i-1}, c)$. For $F_i$, the conditioning vector $c$ replaces the noise vector $z$ to encourage the generators to draw images with more details according to the conditioning variables. Consequently, multi-scale samples are now generated by $s_{i} = G_{i}(h_{i})$. The objective function of training the discriminator $D_i$ for conditional StackGAN-v2 now consists of two terms, the unconditional loss and the conditional loss, 
\begin{equation}\label{eq:hybrid-LD}
\scriptsize
\begin{aligned}
\mathcal{L}_{D_{i}} &= \underbrace{- \mathbb{E}_{x_{i} \sim {p_{data_i}}} \left[\log D_{i}(x_{i})\right] \; - \; \mathbb{E}_{s_{i} \sim {p_{G_{i}}}} \left[\log(1 - D_{i}(s_{i})\right] }_\text{unconditional loss} \; + \\   
&\;\;\;\; \underbrace{- \mathbb{E}_{x_{i} \sim {p_{data_i}}} \left[\log D_{i}(x_{i}, c)\right] \; - \; \mathbb{E}_{s_{i} \sim {p_{G_{i}}}} \left[\log(1 - D_{i}(s_{i}, c)\right] }_\text{conditional loss}.
\end{aligned}
\end{equation}

The unconditional loss determines whether the image is real or fake while the conditional one determines whether the image and the condition match or not. Accordingly, the loss function for each generator $G_i$ is converted to 
\begin{equation}\label{eq:hybrid-LGi}
\begin{aligned}
\mathcal{L}_{G_{i}} &= \underbrace{- \mathbb{E}_{s_{i} \sim {p_{G_{i}}}} \left[\log D_{i}(s_{i})\right] }_\text{unconditional loss}\; + \; \underbrace{- \mathbb{E}_{s_{i} \sim {p_{G_{i}}}} \left[\log D_{i}(s_{i}, c)\right]}_\text{conditional loss}.
\end{aligned}
\end{equation}
The generator $G_i$ at each scale therefore jointly approximates unconditional and conditional image distributions. The final loss for jointly training generators of conditional StackGAN-v2 is computed by substituting Eq.~(\ref{eq:hybrid-LGi}) into Eq.~(\ref{eq:LG}).

\subsection{Color-consistency regularization}\label{sec:color}

As we increase the image resolution at different generators, the generated images at different scales should share similar basic structure and colors. A color-consistency regularization term is introduced to keep samples generated from the same input at different generators more consistent in color and thus to improve the quality of the generated images.

Let $\bm{x_{k}} = (R, G, B)^{T}$ represent a pixel in a generated image, then the mean and covariance of pixels of the given image can be defined by $\bm{\mu}=\sum_{k}\bm{x_{k}}/N$ and $\bm{\Sigma}=\sum_{k}(\bm{x_{k}}-\bm{\mu})(\bm{x_{k}}-\bm{\mu})^{T}/N$, where $N$ is the number of pixels in the image. The color-consistency regularization term aims at minimizing the differences of $\bm{\mu}$ and $\bm{\sigma}$ between different scales to encourage the consistency, which is defined as
\begin{equation} \label{eq:color_consistency}
\small
\mathcal{L}_{C_i} = \frac{1}{n}\sum_{j=1}^{n}	\left(\lambda_{1}\|\bm{\mu}_{s_{i}^{j}}-\bm{\mu}_{s_{i-1}^{j}}\|_{2}^{2} + \lambda_{2} \|\bm{\Sigma}_{s_{i}^{j}}-\bm{\Sigma}_{s_{i-1}^{j}}\|_{F}^{2}	\right), \;
\end{equation}
where $n$ is the batch size, $\bm{\mu}_{s_{i}^{j}}$ and $\bm{\Sigma}_{s_{i}^{j}}$ are mean and covariance for the $j^{th}$ sample generated by the $i^{th}$ generator. Empirically, we set $\lambda_{1} = 1$ and $\lambda_{2}=5$ by default. For the $j^{th}$ input vector, multi-scale samples $s_{1}^{j}$, $s_{2}^{j}$, ..., $s_{m}^{j}$ are generated from $m$ generators of StackGAN-v2. $\mathcal{L}_{C_i}$ can be added to the loss function of the $i^{th}$ generator defined in Eq.~(\ref{eq:LG}) or Eq.~(\ref{eq:hybrid-LGi}), where $i=2,3,...,m$. 
Therefore, the final loss for training the $i^{th}$ generator is defined as ${\mathcal{L}^{\prime}_{G_i}} = \mathcal{L}_{G_i} + \alpha*\mathcal{L}_{C_i}$. 
Experimental results indicate that the color-consistency regularization is very important (\emph{e.g}., $\alpha = 50.0$ in this paper) for the unconditional task, while it  is  not  needed ($\alpha = 0.0$) for the text-to-image synthesis task which has a stronger constraint, \emph{i.e}., the instance-wise correspondence between images and text descriptions. 

\begin{table*}[bt]
\begin{center}
\normalfont
\begin{tabular}{|l|c|c|c|c|c|c|c|c|c|c|}
\hline
 \multirow{2}{4em}{Dataset}&\multicolumn{2}{c|}{CUB~\cite{WahCUB_200_2011}}
 &\multicolumn{2}{c|}{Oxford-102~\cite{Nilsback08}}
 &\multicolumn{2}{c|}{COCO~\cite{LinMBHPRDZ14}} &\multicolumn{2}{c|}{LSUN~\cite{yu15lsun}} &\multicolumn{2}{c|}{ImageNet~\cite{ILSVRC15}} \\
\cline{2-11}
&train &test  &train &test &train &test &bedroom &church &dog &cat \\
\hline
\#Samples &8,855 &2,933 &7,034 &1,155 &80,000 &40,000 &3,033,042 &126,227 &147,873 &6,500 \\
\hline
\end{tabular}
\end{center}
\vspace{-8pt}
    \caption{Statistics of datasets. We do not split LSUN or ImageNet because they are utilized for the unconditional tasks.}
% \vspace{-10pt}
\label{tab:dataset} 
\end{table*}
\begin{table*}[bt]
\begin{center}
\normalfont
\begin{tabular}{|l|c|c|c|c|c|c|c|}
\hline
\multirow{2}{2.8em}{Metric} & \multicolumn{3}{c|}{CUB} &\multicolumn{2}{c|}{Oxford}  &\multicolumn{2}{c|}{COCO}\\
\cline{2-8}
& GAN-INT-CLS & GAWWN & Our StackGAN-v1 
& GAN-INT-CLS & Our StackGAN-v1 &
GAN-INT-CLS & Our StackGAN-v1 \\
\hline
FID~$\downarrow$ &68.79 &67.22 &\bf51.89 &79.55 &\bf55.28 &\bf60.62 &74.05 \\
\hline
FID*~$\downarrow$ &68.79 &53.51 &\bf35.11 &79.55 &\bf43.02 &60.62 &\bf33.88 \\
\hline
IS~$\uparrow$  &2.88~$\pm$~.04 &3.62~$\pm$~.07 &\bf3.70~$\pm$~.04
    &2.66~$\pm$~.03 &\bf3.20~$\pm$~.01
    &7.88~$\pm$~.07 &\bf8.45~$\pm$~.03\\
\hline
IS*~$\uparrow$ &2.88~$\pm$~.04 &\bf3.10~$\pm$~.03 &3.02~$\pm$~.03 
     &2.66~$\pm$~.03 &\bf2.73~$\pm$~.03
     &7.88~$\pm$~.07 &\bf8.35~$\pm$~.11\\
\hline
HR~$\downarrow$  &2.76~$\pm$~.01 &1.95~$\pm$~.02 &\bf1.29~$\pm$~.02
&1.84~$\pm$~.02 &\bf1.16~$\pm$~.02
&1.82~$\pm$~.03 &\bf1.18~$\pm$~.03\\
\hline
\end{tabular}
\end{center}
\vspace{-8pt}
    \caption{Inception scores (IS), fr\'echet inception distance (FID) and average human ranks (HR) of GAN-INT-CLS~\cite{reed2016generative}, GAWWN~\cite{reed2016learning} and our StackGAN-v1 on CUB, Oxford-102, and COCO. (* means that images are re-sized to 64$\times$64 before computing IS* and FID*)}
\vspace{+10pt}
\label{tab:cmp_previous}
\begin{center}
\normalfont
\begin{tabular}{|c|c|c|c|c|c|c|c|c|}
\hline
 \multicolumn{2}{|c|}{Dataset} &CUB &Oxford-102 &COCO &LSUN-bedroom &LSUN-church &ImageNet-dog &ImageNet-cat \\
\hline
\multirow{2}{5em}{FID~$\downarrow$} &StackGAN-v1 &51.89 &55.28 &\bf74.05 &91.94 &57.20 &89.21 &58.73\\ 
\cline{2-9}
&StackGAN-v2 &\bf15.30 &\bf48.68 &81.59 &\bf35.61 &\bf25.36 &\bf44.54 &\bf28.59\\ 
\hline
\multirow{2}{5em}{IS~$\uparrow$} &StackGAN-v1 &3.70~$\pm$~.04 &3.20~$\pm$~.01 &\bf8.45~$\pm$~.03 &\bf3.59~$\pm$~.05 &\bf2.87~$\pm$~.05 &8.84~$\pm$~.08 &\bf4.77~$\pm$~.06\\ 
\cline{2-9}
&StackGAN-v2 &\bf4.04~$\pm$~.05 &\bf3.26~$\pm$~.01 &8.30~$\pm$~.10 &3.02~$\pm$~.04 &2.38~$\pm$~.03 &\bf9.55~$\pm$~.11 &4.23~$\pm$~.05\\ 
\hline
\multirow{2}{5em}{HR~$\downarrow$} &StackGAN-v1 &1.81~$\pm$~.02 &1.70~$\pm$~.03 &\bf1.45~$\pm$~.04 &1.95~$\pm$~.01 &1.86~$\pm$~.02 &1.90~$\pm$~.01 &1.88~$\pm$~.02\\ 
\cline{2-9}
&StackGAN-v2 &\bf1.19~$\pm$~.02 &\bf1.30~$\pm$~.03 &1.55~$\pm$~.05 &\bf1.05~$\pm$~.01 &\bf1.14~$\pm$~.02 &\bf1.10~$\pm$~.01 &\bf1.12~$\pm$~.02 \\ 
\hline
\end{tabular}
\end{center}
\vspace{-8pt}
    \caption{Comparison of StackGAN-v1 and StackGAN-v2 on different datasets by inception scores (IS), fr\'echet inception distance (FID) and average human ranks (HR).}
\label{tab:cmp_v1_v2} 
\end{table*}
\begin{figure*}[tb]
\begin{center}
\includegraphics[width=1.0\linewidth]{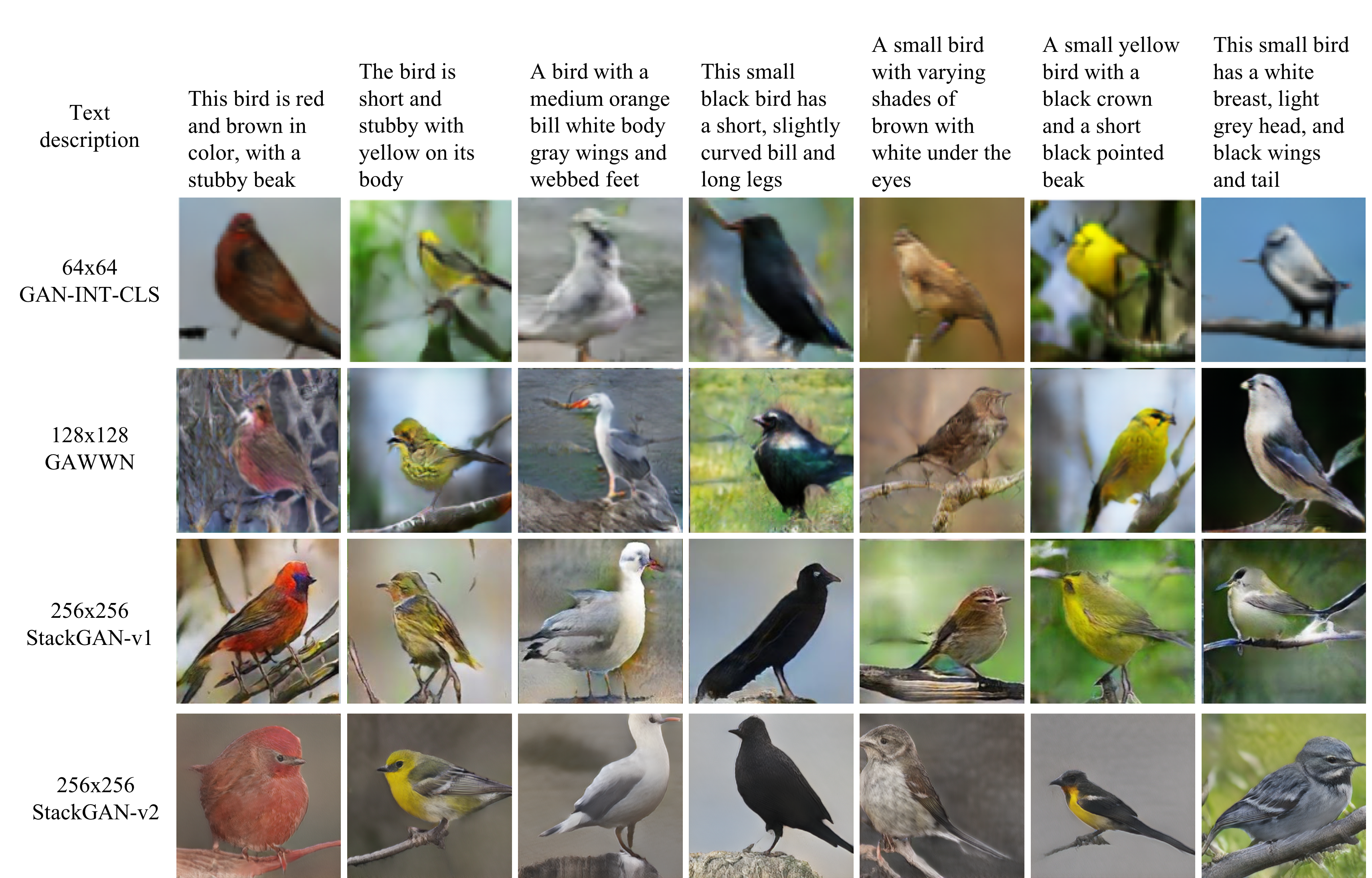}
\end{center}
\vspace{-8pt}
    \caption{Example results by our StackGANs, GAWWN~\cite{reed2016learning}, and GAN-INT-CLS~\cite{reed2016generative} conditioned on text descriptions from CUB test set.}
\vspace{-2pt}
\label{fig:cmp_previous}
\end{figure*}
\begin{figure*}[tb]
\begin{center}
\includegraphics[width=1.0\linewidth]{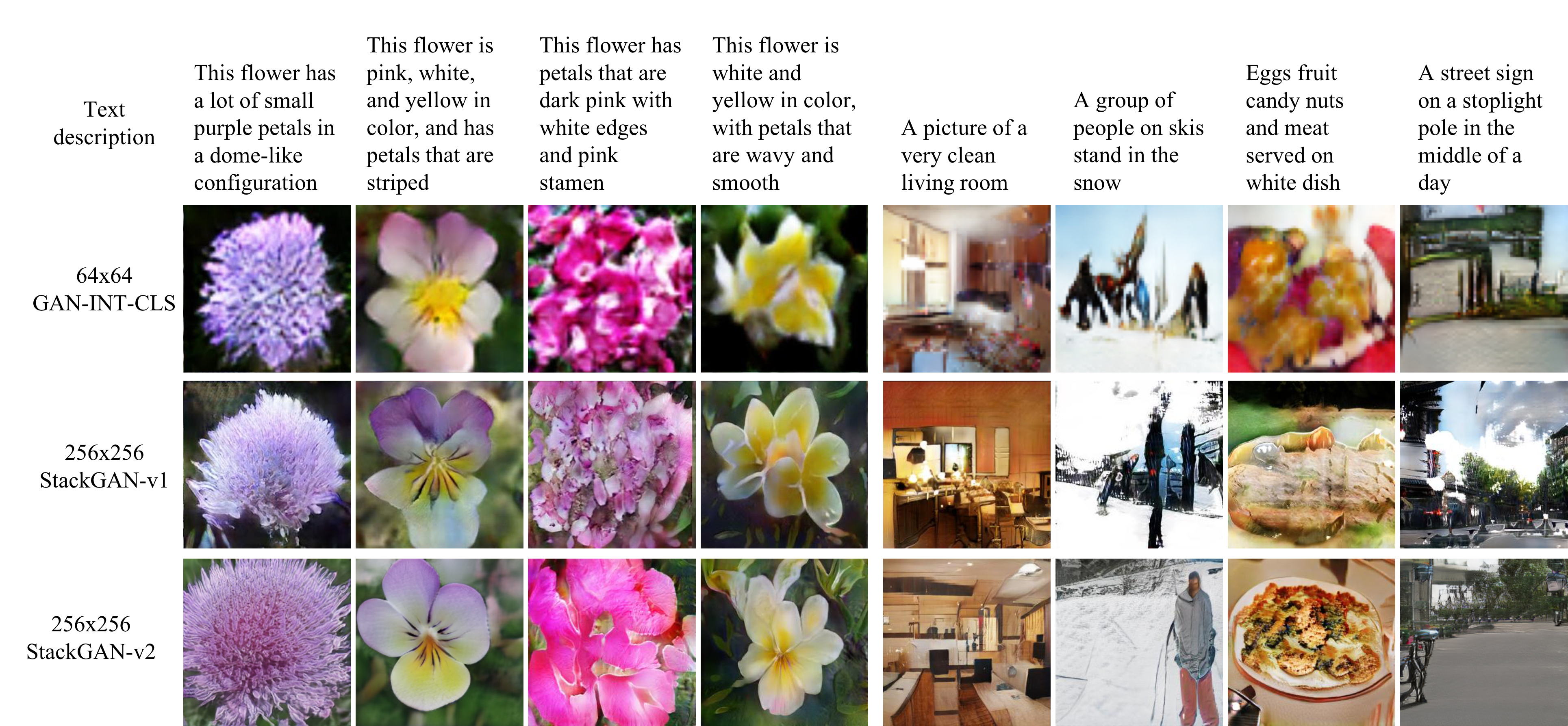}
\end{center}
\vspace{-8pt}
    \caption{Example results by our StackGANs and GAN-INT-CLS~\cite{reed2016generative} conditioned on text descriptions from Oxford-102 test set (leftmost four columns) and COCO validation set (rightmost four columns).}
 \vspace{-8pt}
\label{fig:cmp_previous_flower}
\end{figure*}

\subsection{Implementation details}

As shown in Fig.~\ref{fig:StackGAN-v2}, our StackGAN-v2 models are designed to generate 256$\times$256 images. 
The input vector (\emph{i.e}., $z$ for unconditional StackGAN-v2, or the concatenated $z$ and $c$\footnote{The conditioning variable $c$ for StackGAN-v2 is also generated by Conditioning Augmentation} for conditional StackGAN-v2) is first transformed to a 4$\times$4$\times$64$N_{g}$ feature tensor, where $N_{g}$ is the number of channels in the tensor. Then, this 4$\times$4$\times$64$N_{g}$ tensor is gradually transformed to 64$\times$64$\times$4$N_{g}$, 128$\times$128$\times$2$N_{g}$, and eventually 256$\times$256$\times$1$N_{g}$ tensors at different layers of the network by six up-sampling blocks. The intermediate 64$\times$64$\times$4$N_{g}$, 128$\times$128$\times$2$N_{g}$, and 256$\times$256$\times$1$N_{g}$ features are used to generate images of corresponding scales with 3$\times$3 convolutions. Conditioning variables $c$ or unconditional variables $z$ are also directly fed into intermediate layers of the network to ensure encoded information in $c$ or $z$ is not omitted. All the discriminators $D_i$ have down-sampling blocks and 3$\times$3 convolutions to transform the input image to a 4$\times$4$\times$8$N_{d}$ tensor, and eventually the sigmoid function is used for outputting probabilities.  For all datasets, we set $N_{g}=32$, $N_{d}=64$ and use two residual blocks between every two generators. ADAM~\cite{KingmaB14} solver with $beta1=0.5$ and a learning rate of 0.0002 is used for all models. The source code for StackGAN-v2 is available at {\href{https://github.com/hanzhanggit/StackGAN-v2}{https://github.com/hanzhanggit/StackGAN-v2}} for more implementation details.
%-------------------------------------------------------------------------

\section{Experiments} \label{sec:exp}
{
We conduct extensive experiments to evaluate the proposed methods. In section~\ref{sec:compare}, several state-of-the-art methods on text-to-image synthesis and on unconditional image synthesis are compared with the proposed methods. We first evaluate the effectiveness of our StackGAN-v1 for text-to-image synthesis by comparing it with GAWWN~\cite{reed2016learning} and GAN-INT-CLS~\cite{reed2016generative}. And then, StackGAN-v2 is compared with StackGAN-v1 on different datasets to show its advantages and limitations. Moreover, StackGAN-v2 as a more general framework also works well on unconditional image synthesis tasks, and on such tasks, it is compared with several state-of-the-art methods~\cite{Radford15,Zhao2016,Martin17WGAN,Mao2016,GulrajaniAADC17}. In section~\ref{sec:v1_exp}, several baseline models are designed to investigate the overall design and important components of our StackGAN-v1. For the first baseline, we directly train Stage-\Rmnum{1} GAN for generating 64$\times$64 and 256$\times$256 images to investigate whether the proposed two-stage stacked structure and the Conditioning Augmentation are beneficial. Then we modify our StackGAN-v1 to generate 128$\times$128 and 256$\times$256 images to investigate whether larger images by our method can result in higher image quality. We also investigate whether inputting text at both stages of StackGAN-v1 is useful. In section~\ref{sec:v2_exp}, experiments are designed to validate important components of our StackGAN-v2, including designs with fewer multi-scale image distributions, the effect of jointly approximating conditional and unconditional distributions, and the effectiveness of the proposed color-consistency regularization.  
}

\textbf{Datasets. }
{ 
We evaluate our conditional StackGAN for text-to-image synthesis on the CUB~\cite{WahCUB_200_2011},  Oxford-102~\cite{Nilsback08} and COCO~\cite{LinMBHPRDZ14} datasets. CUB~\cite{WahCUB_200_2011} contains 200 bird species with 11,788 images. Since 80\% of birds in this dataset have object-image size ratios of less than 0.5~\cite{WahCUB_200_2011}, as a pre-processing step, we crop all images to ensure that bounding boxes of birds have greater-than-0.75 object-image size ratios. Oxford-102~\cite{Nilsback08} contains 8,189 images of flowers from 102 different categories.  To show the generalization capability of our approach, a more challenging dataset, COCO~\cite{LinMBHPRDZ14} is also utilized for evaluation. Different from CUB and Oxford-102, the COCO dataset contains images with multiple objects and various backgrounds. Each image in COCO has 5 descriptions, while 10 descriptions are provided by \cite{reed2016cvpr} for every image in CUB and Oxford-102 datasets. Following the experimental setup in \cite{reed2016generative}, we directly use the training and validation sets provided by COCO, meanwhile we split CUB and Oxford-102 into class-disjoint training and test sets. Our unconditional StackGAN utilizes bedroom and church sub-sets of LSUN~\cite{yu15lsun}, a dog-breed  \footnote{Using the wordNet IDs provided by Vinyals \emph{et al}.,~\cite{VinyalsBLKW16}} and  a cat-breed \footnote{The wordNet IDs for this dataset: 'n02121808', 'n02124075', 'n02123394', 'n02122298', 'n02123159','n02123478', 'n02122725', 'n02123597', 'n02124484', 'n02124157', 'n02122878', 'n02123917', 'n02122510', 'n02124313', 'n02123045', 'n02123242', 'n02122430'.} sub-sets of ImageNet~\cite{ILSVRC15} to synthesize different types of images. The statistics of datasets are presented in TABLE~\ref{tab:dataset}. 
}

\textbf{Evaluation metrics. }
{
It is difficult to evaluate the performance of generative models (\emph{e.g}., GANs). In this paper, we choose inception score (IS)~\cite{Salimans2016} and fr\'echet inception distance (FID)~\cite{HeuselRUNH17} for quantitative evaluation. Inception score (IS)~\cite{Salimans2016} is the first well-known metric for evaluating GANs. $IS = \exp \left(\mathbb{E}_{\bm{x}} D_{KL} \left(p\left(y|\bm{x}\right) \,||\, p\left(y\right)\right)\right),$ where $\bm{x}$ denotes one generated sample, and $y$ is the label predicted by the inception model~\cite{Szegedy2016}\footnote{In our experiments, for fine-grained datasets, CUB and Oxford-102, we fine-tune an inception model for each of them. For other datasets, we directly use the pre-trained inception model.}. The intuition behind this metric is that good models should generate diverse but meaningful images. Therefore, the KL divergence between the marginal distribution $p(y)$ and the conditional distribution $p(y|\bm{x})$ should be large. As suggested in \cite{Salimans2016}, we compute the inception score on a large number of samples (\emph{i.e}., 30k samples randomly generated for the test set) for each model\footnote{The mean and standard derivation inception scores of ten splits are reported.}.

Fr\'echet inception distance (FID)~\cite{HeuselRUNH17} was recently proposed as a metric that considers not only the synthetic data distribution but also how it compares to the real data distribution. It directly measures the distance between the synthetic data distribution $p(.)$ and the real data distribution $p_r(.)$. In practice, images are encoded with visual features by the inception model. Assuming the feature embeddings follow a multidimensional Gaussian distribution, the synthetic data's Gaussian with mean and covariance $(m, C)$ is obtained from $p(.)$ and the real data's Gaussian with mean and covariance $(m_r, C_r)$ is obtained from $p_r(.)$. The difference between the synthetic and real Gaussians is measured by the Fr\'echet distance, \emph{i.e}., $FID = ||m-m_r||_{2}^{2} + Tr\left(C + C_r - 2(CC_r)^{1/2}\right)$. Lower FID values mean closer distances between synthetic and real data distributions.  To compute the FID score for a unconditional model, 30k samples are randomly generated. To compute the FID score for a text-to-image model, all sentences in the corresponding test set are utilized to generate samples.

To better evaluate the proposed methods, especially to see whether the generated images are well conditioned on the given  text descriptions, we also conduct user studies. We randomly select 50 text descriptions for each class of CUB and Oxford-102 test sets. For COCO dataset, 4k text descriptions are randomly selected from its validation set. For each sentence, 5 images are generated by each model. Given the same text descriptions, 30 users (not including any of the authors) are asked to rank the results by different methods. The average ranks by human users are calculated to evaluate all compared methods.

In addition, we use t-SNE~\cite{t-SNE} embedding method to visualize a large number (\emph{e.g}., 30k on the CUB test set) of high-dimensional images in a two-dimensional map. We observe that t-SNE is a good tool to examine the distribution of synthesized images and identify collapsed modes. 
}

\begin{figure*}[tb]
\centering
\begin{tabular}{c@{\hspace{2mm}}c}
    \stackunder[5pt]{\includegraphics[width=0.97\columnwidth]{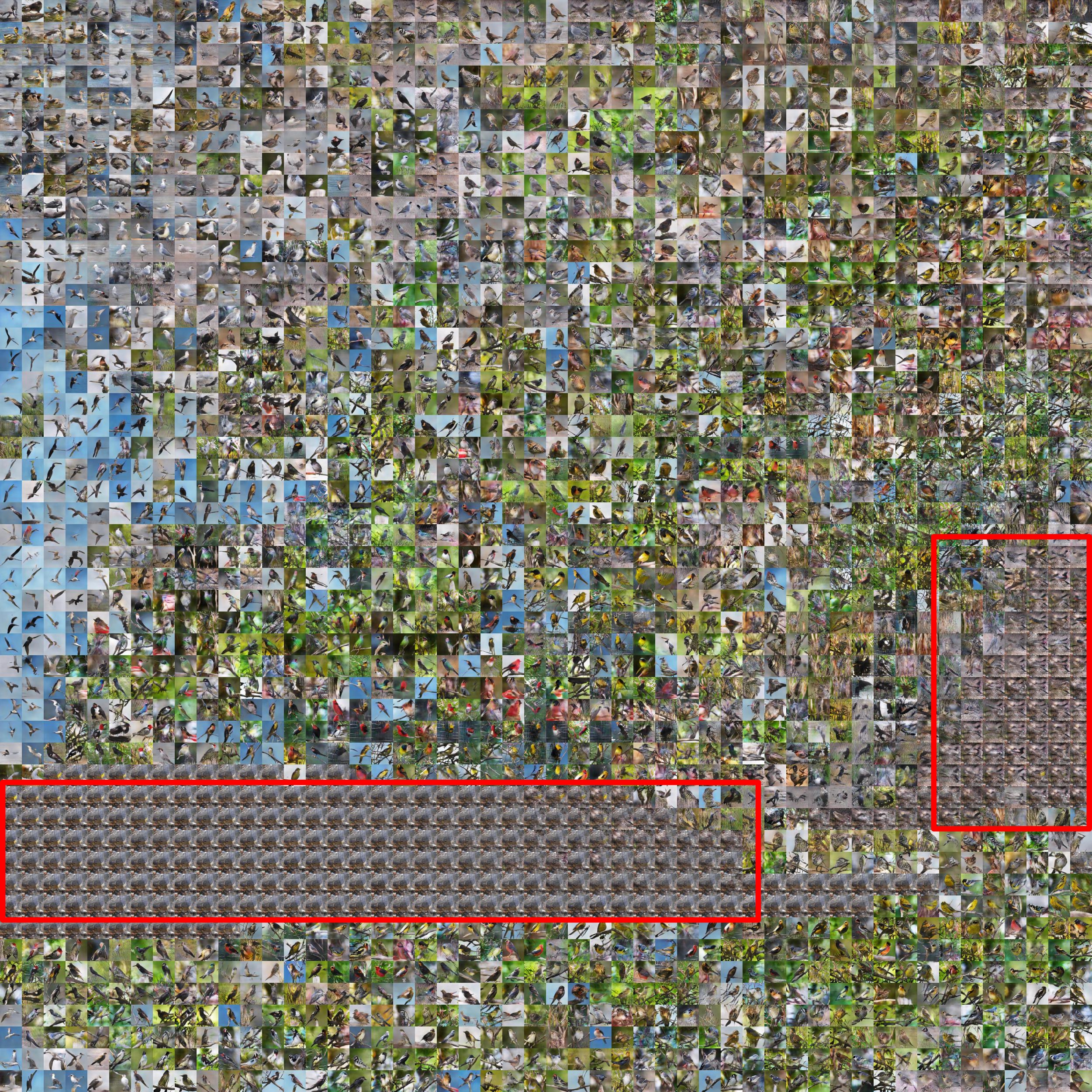}}{(a) StackGAN-v1 has two collapsed modes (in red rectangles).}&
    \stackunder[5pt]{\includegraphics[width=0.97\columnwidth]{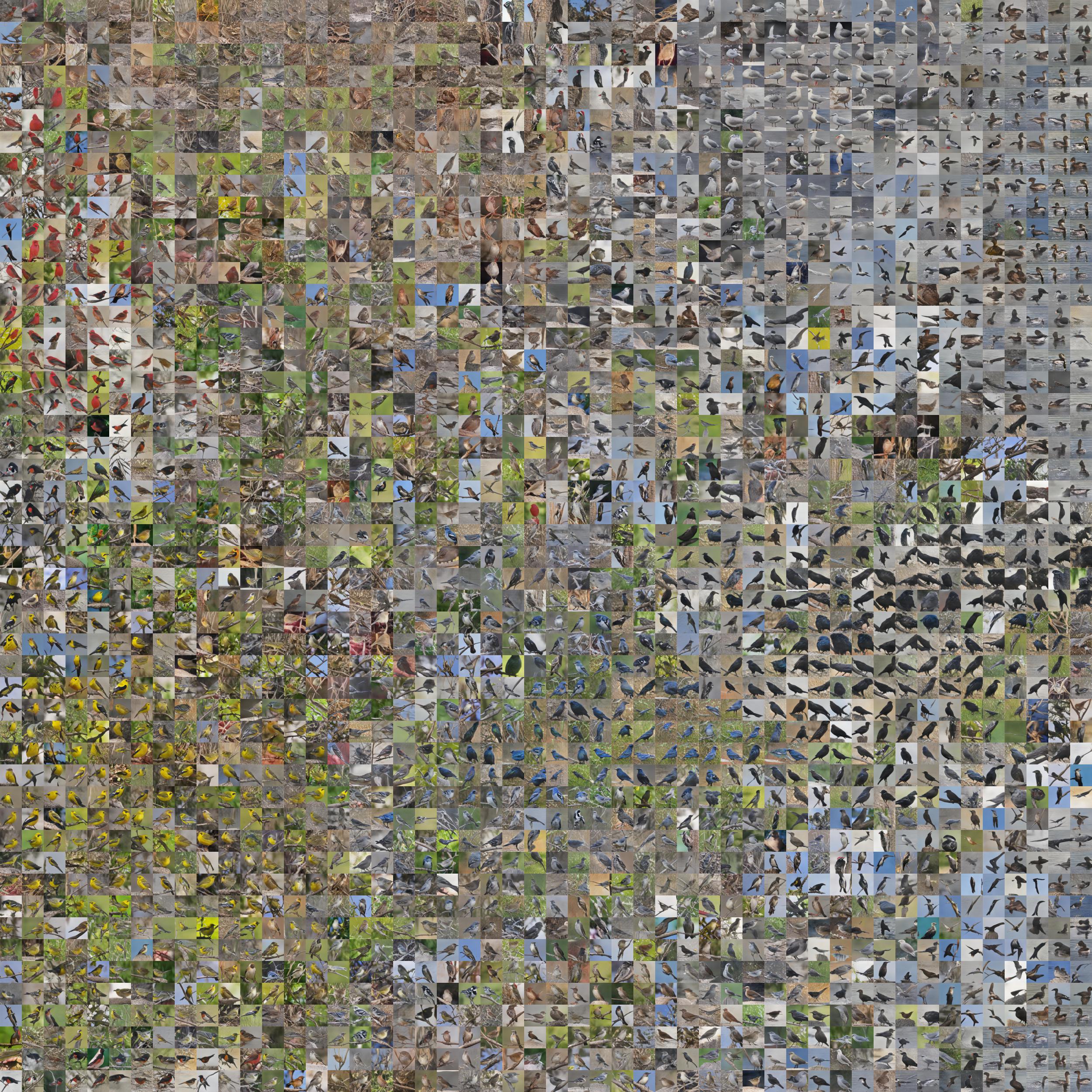}}{(b) StackGAN-v2 contains no collapsed nonsensical mode.}
\end{tabular}
% \vspace{-2pt}
\caption{Utilizing t-SNE to embed images generated by our StackGAN-v1 and StackGAN-v2 on the CUB test set.} 
\label{fig:tsne}
\end{figure*}
\begin{figure*}[tb]
    \centering
    \small
    \begin{tabular}{c@{\hspace{2mm}}c}
    \stackunder[3pt]{\makecell[l]{
        \includegraphics[width=0.935\columnwidth]{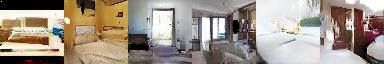}}}{64$\times$64 samples by DCGAN (Reported in~\cite{Radford15})}\vspace{+2pt}&
    \stackunder[3pt]{\makecell[l]{
        \includegraphics[width=0.935\columnwidth]{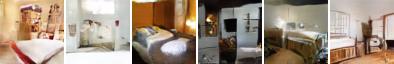}}}{64$\times$64 samples by WGAN (Reported in~\cite{Martin17WGAN})}\vspace{+2pt}\\
    \stackunder[3pt]{\makecell[l]{
        \includegraphics[width=0.935\columnwidth]{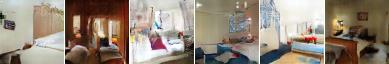}}}{64$\times$64 samples by EBGAN-PT (Reported in~\cite{Zhao2016})}\vspace{+2pt}&
    \stackunder[3pt]{\makecell[l]{
        \includegraphics[width=0.935\columnwidth]{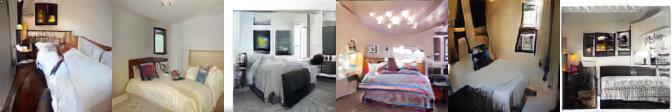}}}{112$\times$112 samples by LSGAN (Reported in~\cite{Mao2016})}\vspace{+2pt}\\
    \stackunder[3pt]{\makecell[l]{
        \includegraphics[width=0.935\columnwidth]{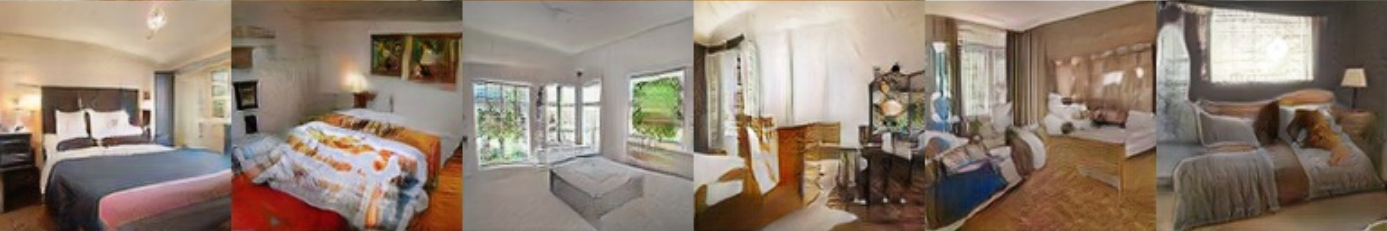}}}{128$\times$128 samples by WGAN-GP (Reported in~\cite{GulrajaniAADC17})}\vspace{+2pt}&
    \stackunder[3pt]{\makecell[l]{
        \includegraphics[width=0.935\columnwidth]{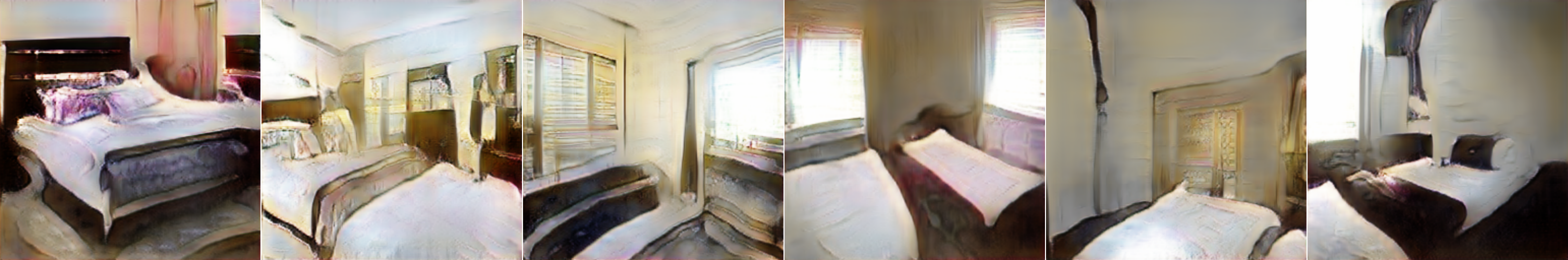}}}{256$\times$256 samples by our StackGAN-v1}\vspace{+2pt}\\
    \end{tabular}
    \begin{tabular}{c}
    \stackunder[3pt]{
        \includegraphics[width=0.23\columnwidth]{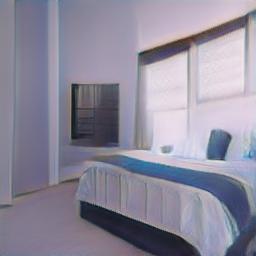}
        \includegraphics[width=0.23\columnwidth]{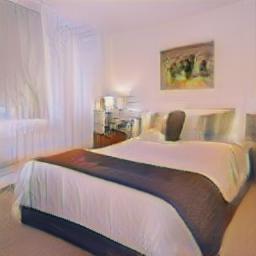}
        \includegraphics[width=0.23\columnwidth]{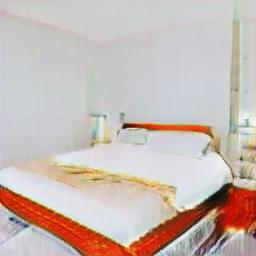}
        \includegraphics[width=0.23\columnwidth]{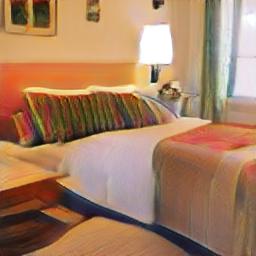}
        \includegraphics[width=0.23\columnwidth]{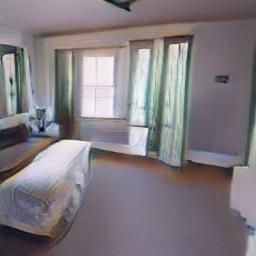}
        \includegraphics[width=0.23\columnwidth]{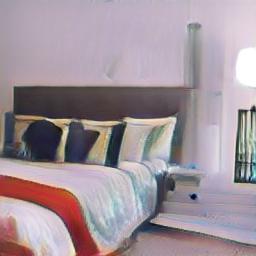}
        \includegraphics[width=0.23\columnwidth]{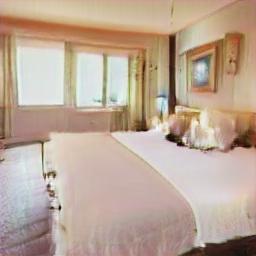}
        \includegraphics[width=0.23\columnwidth]{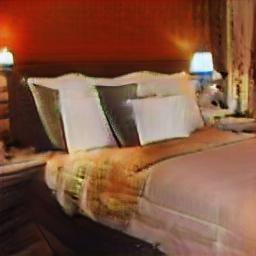}}{256$\times$256 samples by our StackGAN-v2} \\ %\\
    \end{tabular} 
    \vspace{-5pt}
    \caption{Comparison of samples generated by models trained on LSUN bedroom dataset (Zoom in for better comparison).}
\label{fig:compBed}
    \vspace{-10pt}
 \end{figure*}
\begin{figure*}[tb]
    \centering
    \small
    \begin{tabular}{c@{\hspace{2mm}}c}
    \stackunder[3pt]{\makecell[l]{
        \includegraphics[width=0.935\columnwidth]{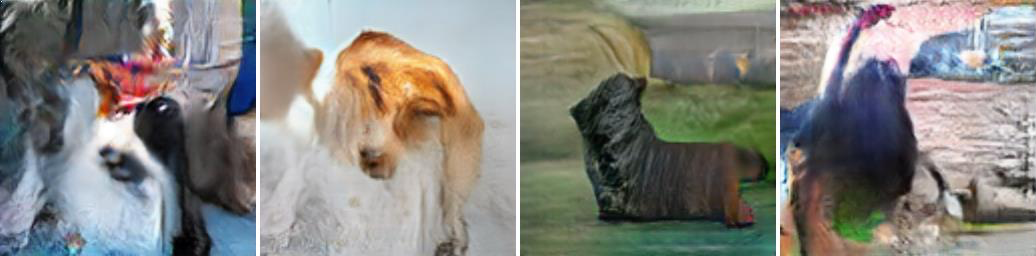}}}{256$\times$256 samples by EBGAN-PT (Reported in~\cite{Zhao2016})}\vspace{+2pt}&
    \stackunder[3pt]{\makecell[l]{
        \includegraphics[width=0.935\columnwidth]{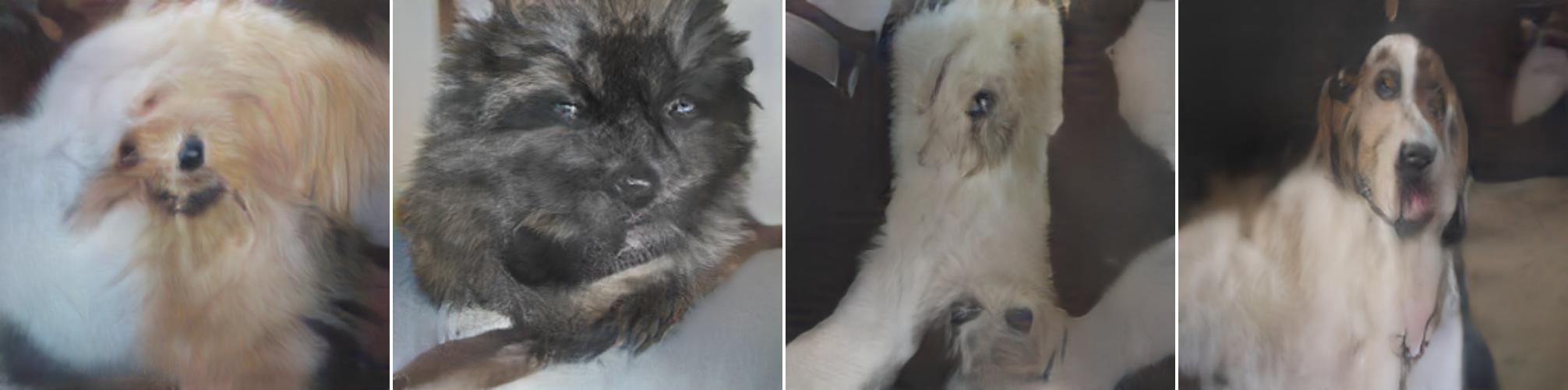}}}{256$\times$256 samples by our StackGAN-v1}\vspace{+2pt}\\
    \end{tabular}
    \begin{tabular}{c}
    \stackunder[3pt]{
        \includegraphics[width=0.23\columnwidth]{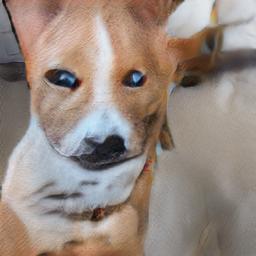}
        \includegraphics[width=0.23\columnwidth]{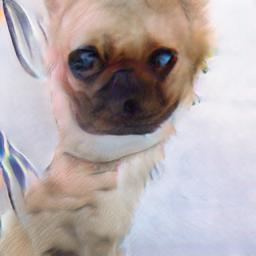}
        \includegraphics[width=0.23\columnwidth]{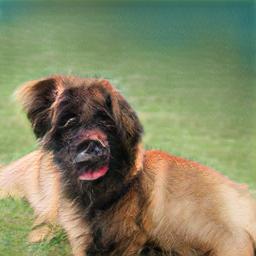}
        \includegraphics[width=0.23\columnwidth]{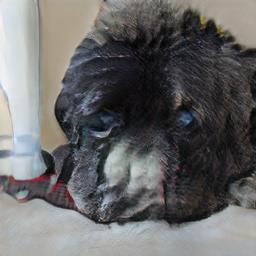}
        \includegraphics[width=0.23\columnwidth]{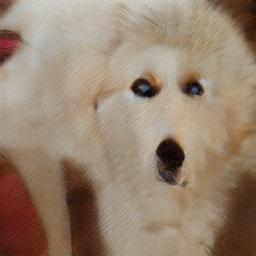}
        \includegraphics[width=0.23\columnwidth]{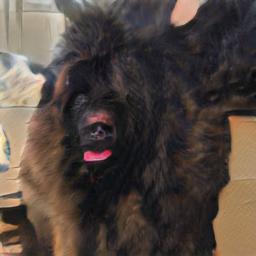}
        \includegraphics[width=0.23\columnwidth]{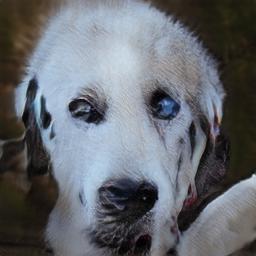}
        \includegraphics[width=0.23\columnwidth]{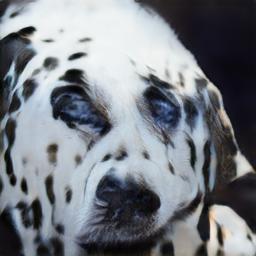}}{256$\times$256 samples by our StackGAN-v2} \\ %\\
    \end{tabular} 
    \vspace{-5pt}
    \caption{Comparison of samples generated by models trained on ImageNet dog dataset.}
    \label{fig:compDog}
    \vspace{+10pt}
%%%%    
    \centering
    \small
    \begin{tabular}{c@{\hspace{2mm}}c}
    \stackunder[3pt]{\makecell[l]{
        \includegraphics[width=0.935\columnwidth]{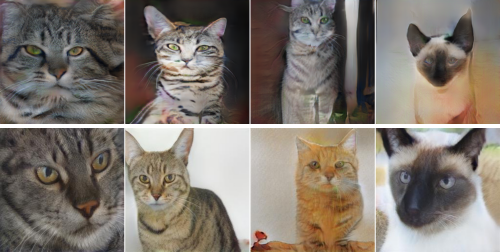}}}{}\vspace{+2pt}&
    \stackunder[3pt]{\makecell[l]{
        \includegraphics[width=0.935\columnwidth]{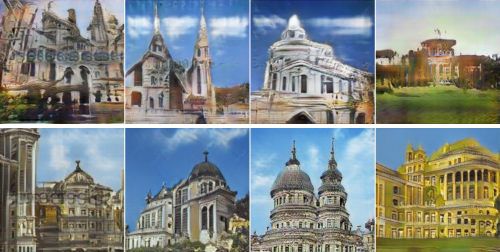}}}{}\vspace{+2pt}\\
    \end{tabular}
    \vspace{-12pt}
    \caption{256$\times$256 samples generated by our StackGAN-v1 (top) and StackGAN-v2 (bottom) on ImageNet cat (left) and LSUN church (right).}
    \label{fig:cat}
    \vspace{-2pt}
 \end{figure*}
\begin{figure*}[tb]
\begin{center}
\includegraphics[width=1.0\linewidth]{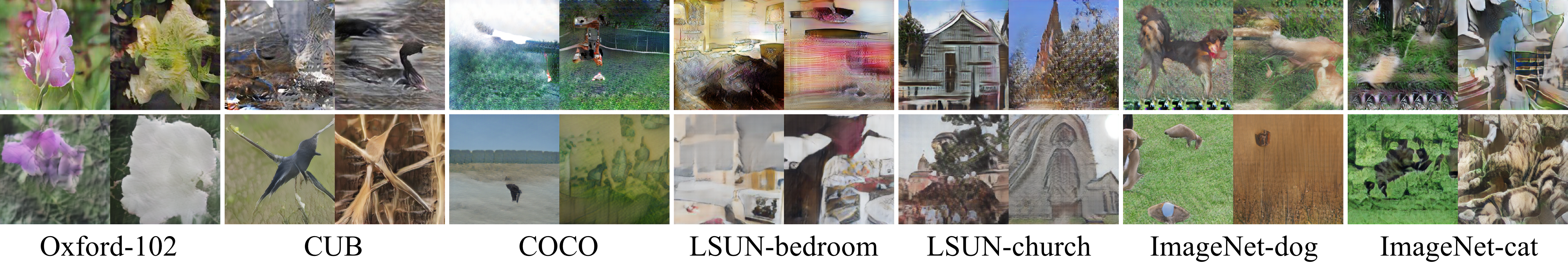}
\end{center}
\vspace{-15pt}
    \caption{Examples of failure cases of StackGAN-v1 (top) and StackGAN-v2 (bottom) on different datasets.}
 \vspace{-12pt}
\label{fig:cmp_failures}
\end{figure*}

\subsection{Comparison with state-of-the-art GAN models}\label{sec:compare} 
To demonstrate the effectiveness of the proposed method, we compare it with state-of-the-art GAN models on text-to-image synthesis~\cite{reed2016generative,reed2016learning} and unconditional image synthesis~\cite{Radford15,Zhao2016,Martin17WGAN,Mao2016, GulrajaniAADC17}.

\textbf{Text-to-image synthesis.}
{
We compare our StackGAN models with several state-of-the-art text-to-image methods~\cite{reed2016learning, reed2016generative} on CUB, Oxford-102 and COCO datasets. The inception scores, fr\'echet inception distances and average human ranks for the proposed StackGAN models and compared methods are reported in TABLE~\ref{tab:cmp_previous}. Representative examples are compared in Fig.~\ref{fig:cmp_previous}, Fig.~\ref{fig:cmp_previous_flower}. For meaningful and fair comparisons with previous methods, the inception scores (IS/IS*) and fr\'echet inception distances (FID/FID*) are computed in two settings. In the first setting, 256$\times$256 images produced by StackGAN, 128$\times$128 images generated by GAWWN~\cite{reed2016learning} and 64$\times$64 images yielded by GAN-INT-CLS~\cite{reed2016generative} are used directly to compute IS and FID. Thus, in this setting, the different models are compared directly using their generated images, which have different resolutions. In the second setting, before computing IS* and FID*, all generated images are re-sized to the same resolution of 64$\times$64 for fair comparison.

Compared with previous GAN models~\cite{reed2016generative,reed2016learning}, on the text-to-image synthesis task, our StackGAN-v1 model achieves the best FID*, IS and average human rank on all three datasets. As shown in TABLE~\ref{tab:cmp_previous}, compared with GAN-INT-CLS~\cite{reed2016generative}, StackGAN-v1 achieves 28.47\% improvement in terms of inception score (IS) on CUB dataset (from 2.88 to 3.70), and 20.30\% improvement on Oxford-102 (from 2.66 to 3.20). When we compare images of different models at the same resolution of 64$\times$64, our StackGAN-v1 still achieves higher inception scores (IS*) than GAN-INT-CLS, but produces a slightly worse inception score (IS*) than GAAWN~\cite{reed2016learning} because GAWWN uses additional supervision. Meanwhile, the FID* of StackGAN-v1 is nearly one half of the FID* of GAN-INT-CLS on each dataset. It means that the StackGAN-v1 can better model and estimate the 64$\times$64 image distribution.  As comparison, the FID of StackGAN-v1 is higher than that of GAN-INT-CLS~\cite{reed2016generative} on COCO. The reason is that the FID of GAN-INT-CLS is the distance between two 64$\times$64 image distributions while the FID of StackGAN-v1 is the distance between two 256$\times$256 image distributions. It is clear that estimating the 64$\times$64 image distribution is much easier than estimating the  256$\times$256 image distribution. It is also the reason why the FID is higher than the FID* for the same model. Finally, the better average human rank of our StackGAN-v1 also indicates our proposed method is able to generate more realistic samples conditioned on text descriptions. On the other hand, representative examples are shown in Fig.~\ref{fig:cmp_previous} and Fig.~\ref{fig:cmp_previous_flower} for visualization comparison. As shown in Fig.~\ref{fig:cmp_previous}, the 64$\times$64 samples generated by GAN-INT-CLS~\cite{reed2016generative} can only reflect the general shape and color of the birds. Their results lack vivid parts (\emph{e.g}., beak and legs) and convincing details in most cases, which make them neither realistic enough nor have sufficiently high resolution. By using additional conditioning variables on location constraints, GAWWN~\cite{reed2016learning} obtains a better inception score on CUB dataset, which is still slightly lower than ours. It generates higher resolution images with more details than GAN-INT-CLS, as shown in Fig.~\ref{fig:cmp_previous}. However, as mentioned by its authors, GAWWN fails to generate any plausible images when it is only conditioned on text descriptions~\cite{reed2016learning}. In comparison, our StackGAN-v1 for the first time generates images of 256$\times$256 resolution with photo-realistic details from only text descriptions. 
}

\textbf{Comparison between StackGAN-v1 and StackGAN-v2. } 
{
The comparison between StackGAN-v1 and StackGAN-v2 by different quantitative metrics as well as human evaluations are reported in TABLE~\ref{tab:cmp_v1_v2}. For unconditional generation, the samples generated by StackGAN-v2 are consistently better than those by StackGAN-v1 (last four columns in TABLE~3) from a human perspective. The end-to-end training scheme together with the color-consistency regularization enables StackGAN-v2 to produce more feedback and regularization for each branch so that consistency is better maintained during the multi-step generation process.  This is especially useful for unconditional generation as no extra conditions (\emph{e.g}., text) are applied.  On the text-to-image datasets, the scores are mixed for StackGAN-v1 and StackGAN-v2. The reason is partially due to the fact that the text information, which is a strong constraint, is added in all the stages to keep coherence. The comparison results of FIDs are consistent with the comparison results of human ranks on all datasets. On the other hand, the inception score draws different conclusions on LSUN-bedroom, LSUN-church, and ImageNet-cat. We think that the reason is because the inception model is pre-trained on ImageNet with 1000 classes, which makes it less suitable for class-specific datasets.  Compared with ImageNet-cat which has 17 classes, the inception score for ImageNet-dog correlates better with human ranks because ImageNet-dog covers more (i.e. 118) classes from ImageNet. Hence we believe that, using class-specific datasets, it is more reasonable to use FID to directly compare feature distances between generated samples with that of the real world samples~\cite{HeuselRUNH17}.

For visual comparison of the results by the two models, we utilize the t-SNE~\cite{t-SNE} algorithm. For each model, a large number of images are generated and embedded into the 2D plane. We first extract a 2048d CNN feature from each generated image using a pre-trained Inception model. Then, t-SNE algorithm is applied to embed the CNN features into a 2D plane, resulting a location for each image in the 2D plane. Due to page limits, Fig.~\ref{fig:tsne} only shows a 50$\times$50 grid with compressed images for each dataset, where each generated image is mapped to its nearest grid location. By visualizing a large number of images, the t-SNE is a good tool to examine the synthesized distribution and evaluate its diversity. We also follow \cite{Odena2016} to use the multiscale structural similarity (MS-SSIM)~\cite{WangSB03} as a metric to measure the variability of samples. We observe that the MS-SSIM is useful to find large-scale mode collapses but often fails to detect small-scale mode collapses or fails to measure the loss of variation in the generated samples' color or texture. This observation is consistent with the one found in ~\cite{KarrasALL18}. For example, in Fig~\ref{fig:tsne}, StackGAN-v1 has two small collapsed modes (nonsensical images) while StackGAN-v2 does not have any collapsed nonsensical mode.  However, the MS-SSIM score of StackGAN-v1 (0.0945) is better than that of StackGAN-v2 (0.1311) and even better than that of the real data (0.1007). Thus, we argue that the MS-SSIM is not a good metric to capture small-scale mode collapses. On the contrary, the t-SNE visualization of the generated samples can easily help us identify any collapsed modes in the samples as well as evaluate sample variability in texture, color and viewpoint.

More visual comparison of StackGAN-v1 and StackGAN-v2 on different datasets can be found in Fig~\ref{fig:cmp_previous}, Fig~\ref{fig:cmp_previous_flower}, Fig~\ref{fig:compBed}, Fig~\ref{fig:compDog}, Fig~\ref{fig:cat}, and Fig.~\ref{fig:cmp_failures}. Specially, Fig.~\ref{fig:cmp_failures} illustrates failure cases of StackGAN-v1 and StackGAN-v2. We categorize the failures in these cases into three groups: mild, moderate, and severe. The ``mild'' group means that the generated images have smooth and coherent appearance but lack vivid objects. The ``moderate'' group means that the generated images have obvious artifacts, which usually are signs of mode collapse. The ``severe'' group indicates that the generated images fall into collapsed modes.  Based on such criterion, on the simple dataset, Oxford-102, all failure cases of StackGAN-v1 belong to the ``mild'' group, while on other datasets all three groups of failure cases are observed. As comparison, we observe that all failure cases of StackGAN-v2 belong to the ``mild'' group, meaning StackGAN-v2-generated images have no collapsed nonsensical mode (see Fig.~\ref{fig:tsne}). By jointly optimizing multiple distributions (objectives), StackGAN-v2 shows more stable training behavior and results in better FID and inception scores on most datasets (see TABLE~\ref{tab:cmp_v1_v2}). However, because of the same reason, compared with StackGAN-v1, it is harder for StackGAN-v2 to converge on more complex datasets, such as COCO.  In contrast, StackGAN-v1 optimizes sub-tasks separately by training stage by stage. It produces slightly more appealing images on COCO than StackGAN-v2 based on human rank results, but also generates more images that are moderate or severe failure cases. Consequently, while StackGAN-v2 is more advanced than StackGAN-v1 in many aspects (such as end-to-end training and more stable training behavior), StackGAN-v1 has the advantage of stage-by-stage training, which converges faster and requires less GPU memory. 
}

\textbf{Unconditional image synthesis.} 
{
We evaluate the effectiveness of StackGAN-v2 for the unconditional image generation task by comparing it with DCGAN~\cite{Radford15}, WGAN~\cite{Martin17WGAN}, EBGAN-PT~\cite{Zhao2016}, LSGAN~\cite{Mao2016}, and WGAN-GP~\cite{GulrajaniAADC17} on the LSUN bedroom dataset. As shown in Fig.~\ref{fig:compBed}, our StackGAN-v2 is able to generate 256$\times$256 images with more photo-realistic details. In Fig.~\ref{fig:compDog}, we also compare the 256$\times$256 samples generated by StackGAN-v2 and EBGAN-PT. As shown in the figure, the samples generated by the two methods have the same resolution, but StackGAN-v2 generates more realistic ones (\emph{e.g}., more recognizable dog faces with eyes and noses). While on LSUN bedroom dataset, only qualitative results are reported in \cite{Radford15,Martin17WGAN,Zhao2016,Mao2016, GulrajaniAADC17}, a DCGAN model~\cite{Radford15} is trained for quantitative comparison using the public available source code~\footnote{https://github.com/carpedm20/DCGAN-tensorflow} on the ImageNet Dog dataset. The inception score of DCGAN is 8.19~$\pm$~0.11 which is much lower than the inception achieved by our StackGAN-v2 (9.55~$\pm$~0.11). These experiments demonstrate that our StackGAN-v2 outperforms the state-of-the-art methods for unconditional image generation.  Example images generated by StackGAN on LSUN church and ImageNet cat datasets are presented in Fig.~\ref{fig:cat}. 
}
%-------------------------------------------------------------------------

%%
\begin{table}
\begin{center}
\normalfont
\begin{tabular}{|c|c|c|c|}
\hline
Method &CA &Text twice  &Inception score \\
\hline
\multirow{2}{3cm}{\,\,\,\,\,64$\times$64 Stage-\Rmnum{1} GAN}
                                  &no    &/     &2.66~$\pm$~.03   \\
                                  &yes   &/     &2.95~$\pm$~.02   \\
\hline
\multirow{2}{3cm}{\,\,\,\,256$\times$256 Stage-\Rmnum{1} GAN}
                                  &no    &/     &2.48~$\pm$~.00   \\
                                  &yes    &/     &3.02~$\pm$~.01   \\
\hline
\multirow{3}{3cm}{\,\,\,\,128$\times$128 StackGAN-v1}
                                  &yes   &no    &3.13~$\pm$~.03  \\ 
                                  &no    &yes   &3.20~$\pm$~.03  \\
                                  &yes   &yes   &3.35~$\pm$~.02  \\
\hline
\multirow{3}{3cm}{\,\,\,\,256$\times$256 StackGAN-v1}
                                  &yes   &no    &3.45~$\pm$~.02  \\ 
                                  &no    &yes   &3.31~$\pm$~.03  \\
                                  &yes   &yes   &3.70~$\pm$~.04  \\
\hline
\end{tabular}
\end{center}
\vspace{-5pt}
    \caption{Inception scores calculated with 30,000 samples generated on CUB by different baseline models of our StackGAN-v1.}
\label{tab:inception_score}
\vspace{-5pt}
\end{table}
\begin{figure*}[bt]
\begin{center}
	\includegraphics[width=1.0\linewidth]{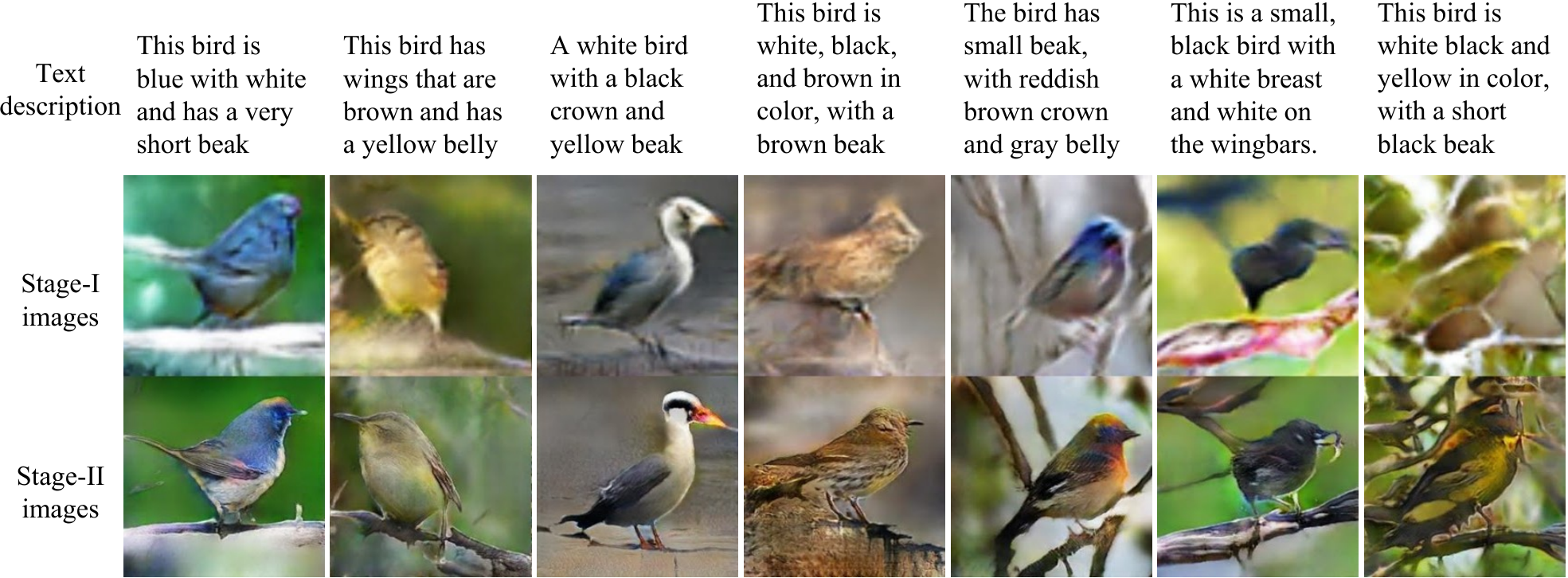}
\end{center}
\vspace{-8pt}
   \caption{Samples generated by our StackGAN-v1 from unseen texts in CUB test set. 
   Each column lists the text description, images generated from the text by Stage-\Rmnum{1} and Stage-\Rmnum{2} of StackGAN-v1.}
\vspace{-5pt}
\label{fig:lr2hr}
\end{figure*}
\begin{figure}[bt]
\begin{center}
	\includegraphics[width=1.0\linewidth]{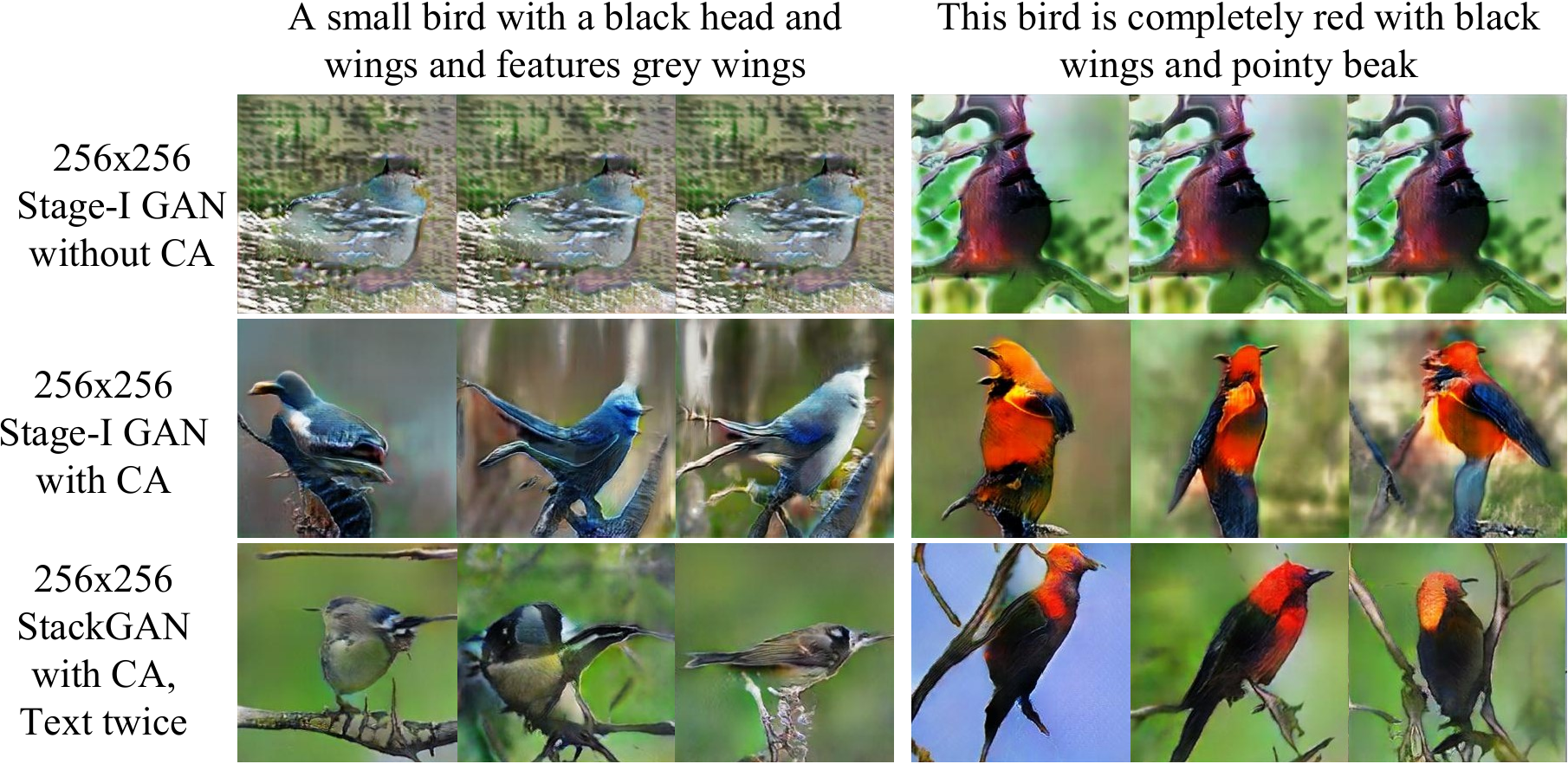}
\end{center}
\vspace{-8pt}
   \caption{Conditioning Augmentation (CA) helps stabilize the training of conditional GAN and improves the diversity of the generated samples. 
   (Row 1) without CA, Stage-\Rmnum{1} GAN fails to generate plausible 256$\times$256 samples. Although different noise vector $z$ is used for each column, the generated samples collapse to be the same for each input text description. 
   (Row 2-3) with CA but fixing the noise vectors $z$, methods are still able to generate birds with different poses and viewpoints.}
\vspace{-5pt}
\label{fig:Gaussian}
\end{figure}
\begin{figure}[bt]
\begin{center}
	\includegraphics[width=1\linewidth]{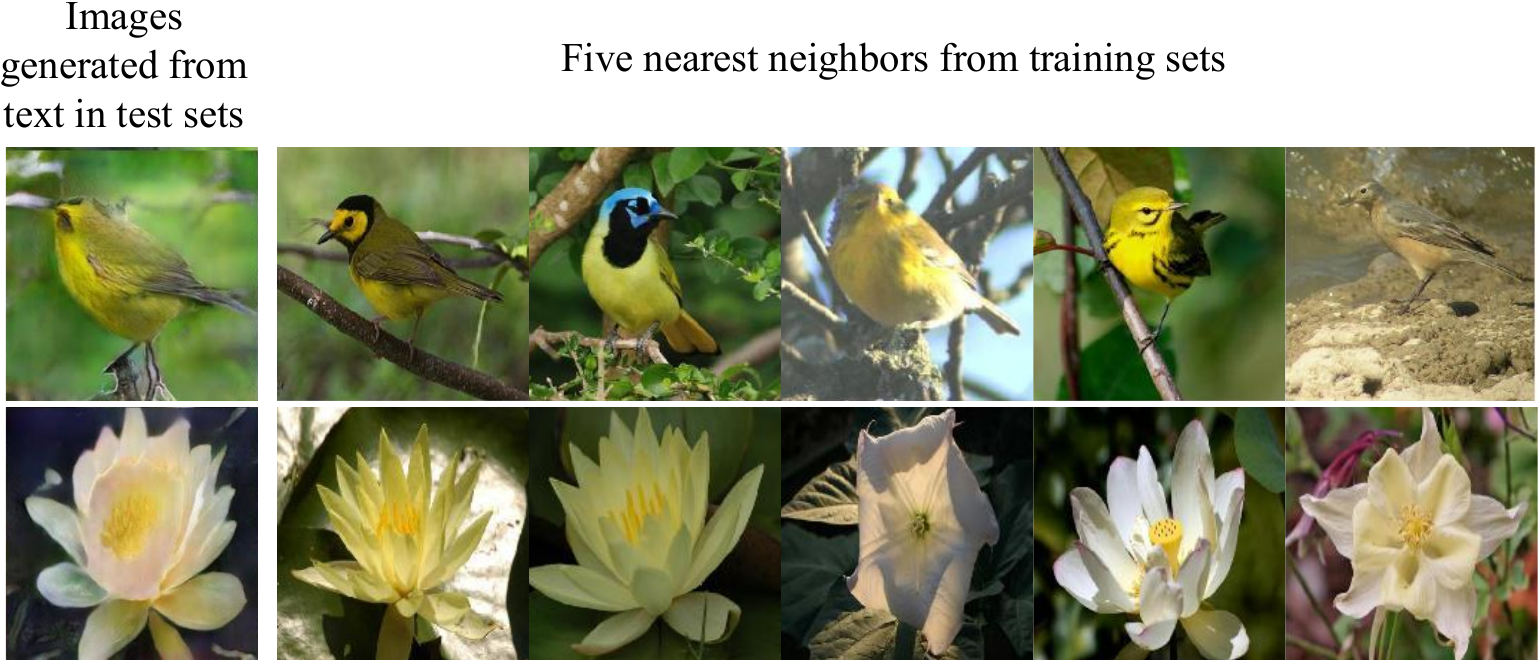}
\end{center}
\vspace{-8pt}
   \caption{For generated images (column 1), retrieving their nearest training images (columns 2-6) by utilizing Stage-II discriminator of StackGAN-v1 to extract visual features. The $L2$ distances between features are calculated for nearest-neighbor retrieval.}
\vspace{-5pt}
\label{fig:NNs}
\end{figure}
\begin{figure}[bt]
\begin{center}
	\includegraphics[width=0.9\linewidth]{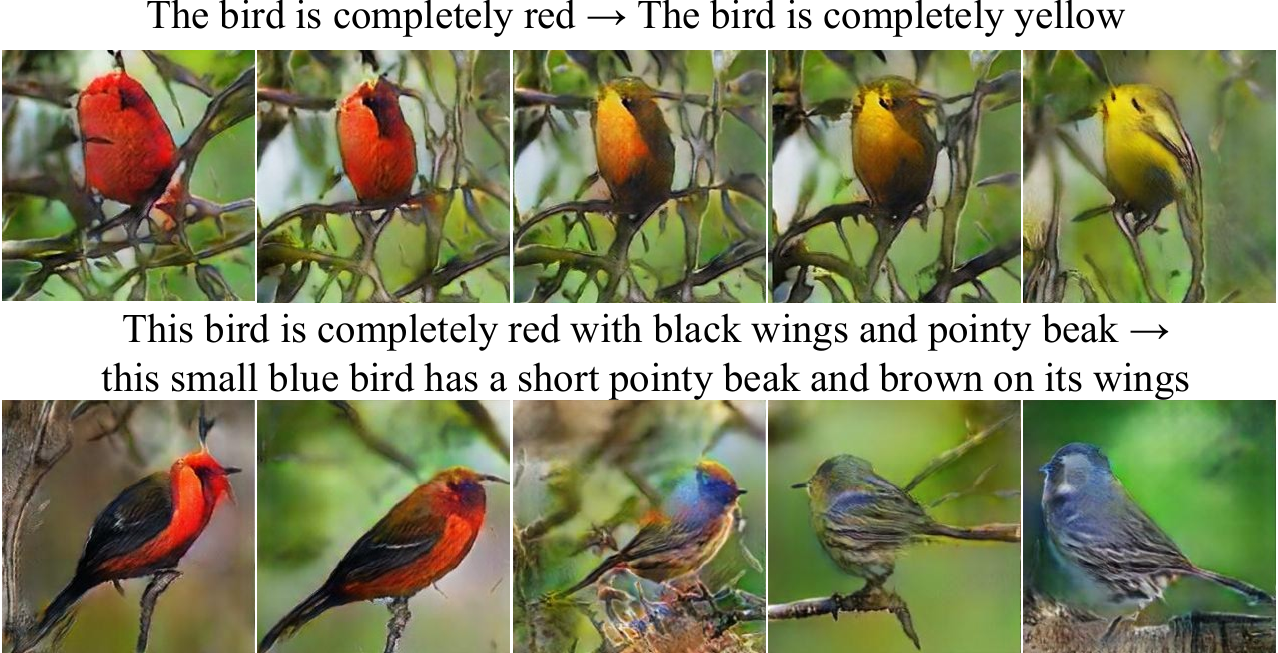}
\end{center}
\vspace{-8pt}
   \caption{(Left to right) Images generated by interpolating two sentence embeddings. Gradual appearance changes from the first sentence's meaning to that of the second sentence can be observed. The noise vector $z$ is fixed to be zeros for each row.}
\vspace{-10pt}
\label{fig:interpolate}
\end{figure}
%%%

\subsection{The component analysis of StackGAN-v1}\label{sec:v1_exp}

In this section, we analyze different components of StackGAN-v1 on CUB dataset with baseline models.  

\textbf{The design of StackGAN-v1.}
{　
As shown in the first four rows of TABLE~\ref{tab:inception_score}, if Stage-I GAN is directly used to generate images, the inception scores decrease significantly. Such performance drop can be well illustrated by results in Fig.~\ref{fig:Gaussian}. As shown in the first row of Fig.~\ref{fig:Gaussian}, Stage-\Rmnum{1} GAN fails to generate any plausible 256$\times$256 samples without using Conditioning Augmentation (CA). Although Stage-\Rmnum{1} GAN with CA is able to generate more diverse 256$\times$256 samples, those samples are not as realistic as samples generated by StackGAN-v1. It demonstrates the necessity of the proposed stacked structure. In addition, by decreasing the output resolution from 256$\times$256 to 128$\times$128, the inception score decreases from 3.70 to 3.35. Note that all images are scaled to 299 $\times$ 299 before calculating the inception score. Thus, if our StackGAN-v1 just increases the image size without adding more information, the inception score would remain the same for samples of different resolutions. Therefore, the decrease in inception score by 128$\times$128 StackGAN-v1 demonstrates that our 256$\times$256 StackGAN-v1 does add more details into the larger images. For the 256$\times$256 StackGAN-v1, if the text is only input to Stage-I (denoted as ``no Text twice''), the inception score decreases from 3.70 to 3.45. It indicates that processing text descriptions again at Stage-II helps refine Stage-I results. The same conclusion can be drawn from the results of 128$\times$128 StackGAN-v1 models.

Fig.~\ref{fig:lr2hr} illustrates some examples of the Stage-I and Stage-II images generated by our StackGAN-v1. As shown in the first row of Fig.~\ref{fig:lr2hr}, in most cases, Stage-\Rmnum{1} GAN is able to draw rough shapes and colors of objects given text descriptions. However, Stage-\Rmnum{1} images are usually blurry with various defects and missing details, especially for foreground objects. As shown in the second row, Stage-\Rmnum{2} GAN generates 4$\times$ higher resolution images with more convincing details to better reflect corresponding text descriptions. For cases where Stage-\Rmnum{1} GAN has generated plausible shapes and colors, Stage-\Rmnum{2} GAN completes the details. For instance, in the $1$st column of Fig.~\ref{fig:lr2hr}, with a satisfactory Stage-\Rmnum{1} result, Stage-\Rmnum{2} GAN focuses on drawing the short beak and white color described in the text as well as details for the tail and legs. In all other examples, different degrees of details are added to Stage-\Rmnum{2} images. In many other cases, Stage-\Rmnum{2} GAN is able to correct the defects of Stage-\Rmnum{1} results by processing the text description again. For example, while the Stage-\Rmnum{1} image in the $5$th column has a blue crown rather than the reddish brown crown described in the text, the defect is corrected by Stage-\Rmnum{2} GAN. In some extreme cases (\emph{e.g}., the $7$th column of Fig.~\ref{fig:lr2hr}), even when Stage-\Rmnum{1} GAN fails to draw a plausible shape, Stage-\Rmnum{2} GAN is able to generate reasonable objects. We also observe that StackGAN-v1 has the ability to transfer background from Stage-\Rmnum{1} images and fine-tune them to be more realistic with higher resolution at Stage-\Rmnum{2}.

Importantly, the StackGAN-v1 does not achieve good results by simply memorizing training samples but by capturing the complex underlying language-image relations. By feeding our generated images and all training images to the Stage-II discriminator $D$ of our StackGAN-v1, their visual features are extracted from the last Conv. layer of $D$. And then, we can compute the similarity between two images, based on their visual features. Finally, for each generated image, its nearest neighbors from the training set can be retrieved. By visually inspecting the retrieved images (see Fig.~\ref{fig:NNs}), we conclude that the generated images have some similar characteristics with the training samples but are essentially different. 
}

\textbf{Conditioning Augmentation.}
{
We also investigate the efficacy of the proposed Conditioning Augmentation (CA). By removing it from StackGAN-v1 256$\times$256 (denoted as ``no CA'' in TABLE~\ref{tab:inception_score}), the inception score decreases from 3.70 to 3.31. Fig.~\ref{fig:Gaussian} also shows that 256$\times$256 Stage-\Rmnum{1} GAN (and StackGAN-v1) with CA can generate birds with different poses and viewpoints from the same text embedding. In contrast, without using CA, samples generated by 256$\times$256 Stage-\Rmnum{1} GAN collapse to nonsensical images due to the unstable training dynamics of GANs. Consequently, the proposed Conditioning Augmentation helps stabilize the conditional GAN training and improves the diversity of the generated samples because of its ability to encourage robustness to small perturbations along the latent manifold. 
}

\textbf{Sentence embedding interpolation. }
{
To further demonstrate that our StackGAN-v1 learns a smooth latent data manifold, we use it to generate images from linearly interpolated sentence embeddings, as shown in Fig.~\ref{fig:interpolate}. We fix the noise vector $z$, so the generated image is inferred from the given text description only. Images in the first row are generated by simple sentences made up by us. Those sentences contain only simple color descriptions. The results show that the generated images from interpolated embeddings can accurately reflect color changes and generate plausible bird shapes. The second row illustrates samples generated from more complex sentences, which contain more details on bird appearances. The generated images change their primary color from red to blue, and change the wing color from black to brown.
}

%% %% %% %% %% %% %% %% %%
%%
\begin{table*}[bt]
\begin{center}
\normalfont
\begin{tabular}{|l|c|c|c|c|c|}
\hline
Model &branch $G_1$ &branch $G_2$ &branch $G_3$  &JCU  & inception score\\
\hline
StackGAN-v2&64$\times$64 &128$\times$128 &256$\times$256   &yes  &4.04~$\pm$~.05\\
\hline
StackGAN-v2-no-JCU     &64$\times$64 &128$\times$128 &256$\times$256  &no   &3.77~$\pm$~.04\\
\hline
StackGAN-v2-$G_3$   &removed &removed &256$\times$256  &yes &3.49~$\pm$~.04\\
\hline
StackGAN-v2-3$G_3$   &removed &removed &three 256$\times$256  &yes &3.22~$\pm$~.02\\
\hline
StackGAN-v2-all256 &256$\times$256 &256$\times$256 &256$\times$256    &yes    &2.89~$\pm$~.02\\
\hline
\end{tabular}
\end{center}
% \vspace{-5pt}
    \caption{Inception scores by our StackGAN-v2 and its baseline models on CUB test set. ``JCU'' means using the proposed discriminator that jointly approximates conditional and unconditional distributions.}
\vspace{-5pt}
\label{tab:baseline} 
\end{table*}
\begin{figure*}[tb]
    \centering
    \small
    \begin{tabular}{c@{\hspace{2mm}}c@{\hspace{2mm}}c@{\hspace{2mm}}c}
    \stackunder[5pt]{
    \includegraphics[width=0.23\columnwidth]{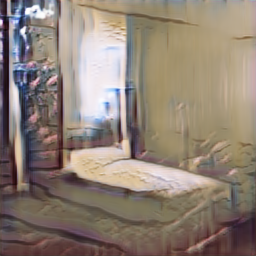}
    \includegraphics[width=0.23\columnwidth]{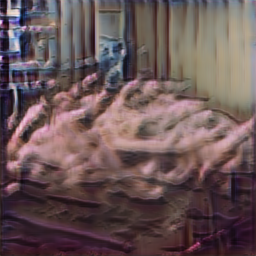}}{(a) StackGAN-v2-all256}& 
    \stackunder[5pt]{
    \includegraphics[width=0.23\columnwidth]{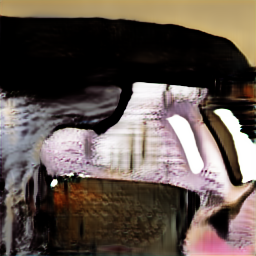}
    \includegraphics[width=0.23\columnwidth]{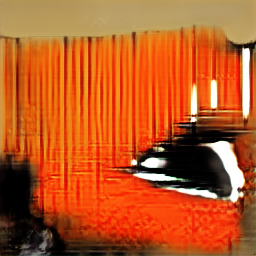}}{(b) StackGAN-v2-$G_3$}&
    \stackunder[5pt]{
    \includegraphics[width=0.23\columnwidth]{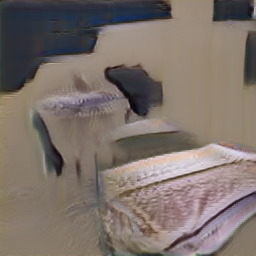}
    \includegraphics[width=0.23\columnwidth]{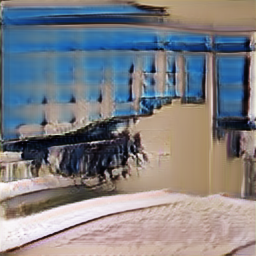}}{(c) StackGAN-v2-3$G_3$}&
    \stackunder[5pt]{
    \includegraphics[width=0.23\columnwidth]{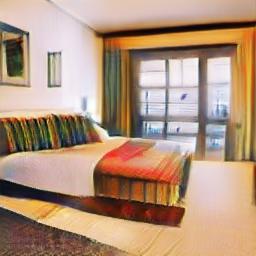}
    \includegraphics[width=0.23\columnwidth]{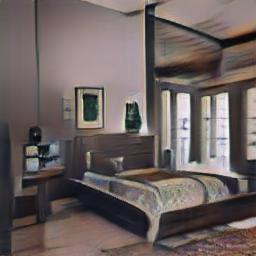}}{(d) StackGAN-v2}\vspace{+8pt} \\
    \multicolumn{4}{c}{This black and white and grey bird has a black bandit marking around it's eyes} \\
    \stackunder[5pt]{
    \includegraphics[width=0.23\columnwidth]{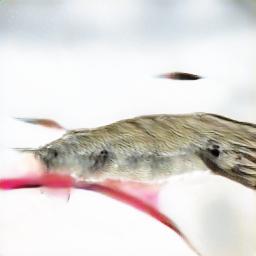}
    \includegraphics[width=0.23\columnwidth]{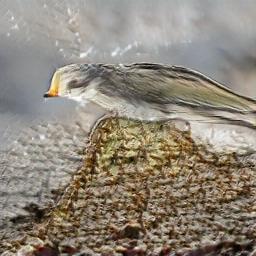}}{(e) StackGAN-v2-all256}& 
    \stackunder[5pt]{
    \includegraphics[width=0.23\columnwidth]{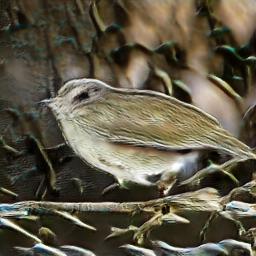}
    \includegraphics[width=0.23\columnwidth]{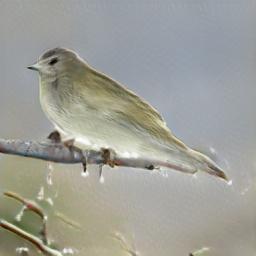}}{(f) StackGAN-v2-$G_3$}& 
    \stackunder[5pt]{
    \includegraphics[width=0.23\columnwidth]{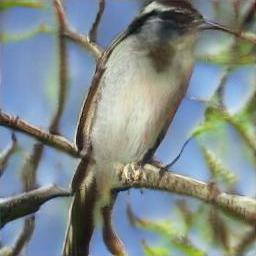} \includegraphics[width=0.23\columnwidth]{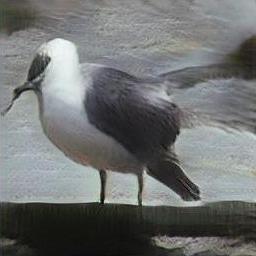}}{(g) StackGAN-v2-no-JCU}& 
    \stackunder[5pt]{
    \includegraphics[width=0.23\columnwidth]{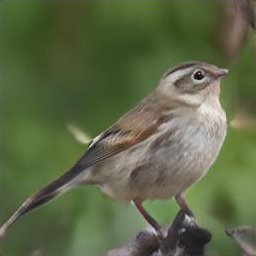}
    \includegraphics[width=0.23\columnwidth]{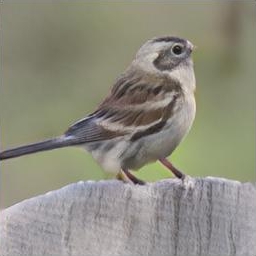}}{(h) StackGAN-v2}
    \end{tabular}
    % \vspace{-4pt}
    \caption{Example images generated by the StackGAN-v2 and its baseline models on LSUN bedroom (top) and CUB (bottom) datasets.} 
    \label{fig:multiD}
    % \vspace{-5pt}
 \end{figure*}
 \begin{figure*}[tb]
    \centering
    \small
    \begin{tabular}{c@{\hspace{4mm}}c@{\hspace{4mm}}c}
    \stackunder[3pt]{\makecell[l]{
     \includegraphics[width=17mm]{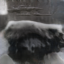}
     \includegraphics[width=17mm]{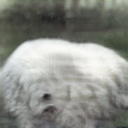}
     \includegraphics[width=17mm]{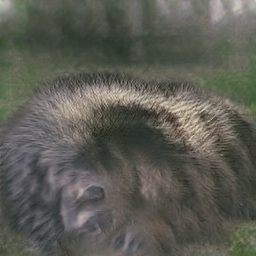}\\
     \includegraphics[width=17mm]{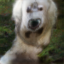}
     \includegraphics[width=17mm]{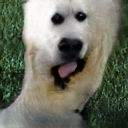}
     \includegraphics[width=17mm]{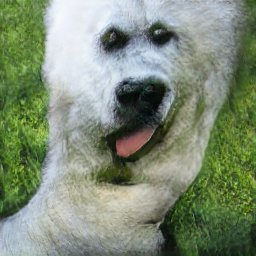}}}{ImageNet dog~\cite{ILSVRC15}}&
    \stackunder[3pt]{\makecell[l]{
    \includegraphics[width=17mm]{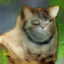}
     \includegraphics[width=17mm]{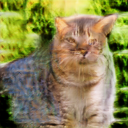}
     \includegraphics[width=17mm]{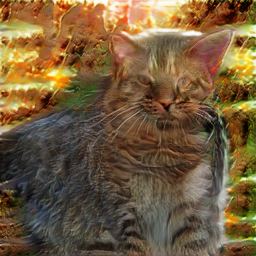}\\
     \includegraphics[width=17mm]{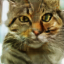}
     \includegraphics[width=17mm]{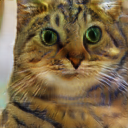}
     \includegraphics[width=17mm]{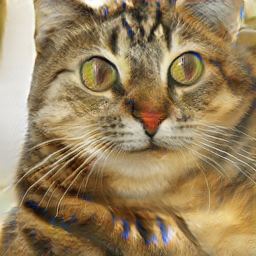}}}{ImageNet cat~\cite{ILSVRC15}}& 
    \stackunder[3pt]{\makecell[l]{
     \includegraphics[width=17mm]{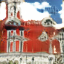}
     \includegraphics[width=17mm]{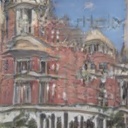}
     \includegraphics[width=17mm]{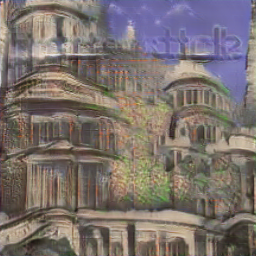}\\
     \includegraphics[width=17mm]{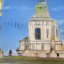}
     \includegraphics[width=17mm]{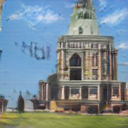}
     \includegraphics[width=17mm]{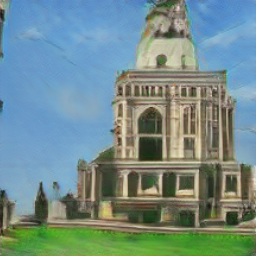}}}{LSUN church~\cite{yu15lsun}}
    \end{tabular}
    % \vspace{-4pt}
    \caption{Example images generated without (top) and with (bottom) the proposed color-consistency regularization for our StackGAN-v2 on ImageNet dog, ImageNet cat and LSUN church datasets. 
    (Left to right) 64$\times$64, 128$\times$128, and 256$\times$256 images by $G_1$, $G_2$, $G_3$, respectively.}
    \vspace{-5pt}
    \label{fig:color}
 \end{figure*}

\subsection{The component analysis of StackGAN-v2}\label{sec:v2_exp}

In this section, we analyze important components of the proposed StackGAN-v2. TABLE~\ref{tab:baseline} lists models with different settings and their inception scores on the CUB test set. Fig.~\ref{fig:multiD} shows example images generated by different baseline models.

Our baseline models are built by removing or changing a certain component of the StackGAN-v2 model. By approximating the image distribution directly at the 256$\times$256 scale without intermediate branches, the inception scores on CUB dramatically decrease from 4.04 to 3.49 for  ``StackGAN-v2-$G_3$'' and to 2.89 for ``StackGAN-v2-all256'' (See TABLE~\ref{tab:baseline} and Figures~\ref{fig:multiD} (e-f)). This demonstrates the importance of the multi-scale, multi-stage architecture in StackGAN-v2. Inspired by~\cite{DurugkarGM17}, we also build a baseline model with multiple discriminators at the 256$\times$256 scale, namely ``StackGAN-v2-3$G_3$''. Those discriminators have the same structure but different initializations. However, the results do not show improvement over ``StackGAN-v2-$G_3$''. Similar comparisons have also been done for the unconditional task on the LSUN bedroom dataset. As shown in Figures~\ref{fig:multiD}(a-c), those baseline models fail to generate realistic images because they suffer from severe mode collapses.

To further demonstrate the effectiveness of jointly approximating conditional and unconditional distributions, ``StackGAN-v2-no-JCU'' replaces the jointly conditional and unconditional discriminators with the conventional ones,  resulting in much lower inception score than that of ``StackGAN-v2''. Another baseline model does not use the color-consistency regularization term.  Results on various datasets (see Fig.~\ref{fig:color}) show that the color-consistency regularization has significant positive effects for the unconditional image synthesis task. Quantitatively, removing the color-consistency regularization decreases the inception score from 9.55~$\pm$~0.11 to 9.02~$\pm$~0.14 on the ImageNet dog dataset. It demonstrates that the additional
constraint provided by the color-consistency regularization is able to facilitate multi-distribution approximation and help generators at different branches produce more coherent samples. It is worth mentioning that there is no need to utilize the color-consistency regularization for the text-to-image synthesis task because the text conditioning appears to provide sufficient constraints. Experimentally, adding the color-consistency regularization did not improve the inception score on CUB dataset.

\section{Conclusions}\label{sec:conclusion}

In this paper, Stacked Generative Adversarial Networks, StackGAN-v1 and StackGAN-v2, are proposed to decompose the difficult problem of generating realistic high-resolution images into more manageable sub-problems. The StackGAN-v1 with Conditioning Augmentation is first proposed for text-to-image synthesis through a novel sketch-refinement process. It succeeds in generating images of 256$\times$256 resolution with photo-realistic details from text descriptions. To further improve the quality of generated samples and stabilize GANs' training, the StackGAN-v2 jointly approximates multiple related distributions, including (1) multi-scale image distributions and (2) jointly conditional and unconditional image distributions. In addition, a color-consistency regularization is proposed to facilitate multi-distribution approximation. Extensive quantitative and qualitative results demonstrate that our proposed methods significantly improve the state of the art in both conditional and unconditional image generation tasks. 

%-------------------------------------------------------------------------

% \section{Acknowledgement}

\bibliographystyle{ieee}
\bibliography{paperbib}

% \section{author bios & photos}
% Comments the title and put author bios & photos here. 
% The author order on the first page and in the bios should be the same. Also, an author’s name on the first page and in the bio should match exactly. 
\vspace{+10pt}
\small
\begin{wrapfigure}{l}{0.2\columnwidth}
    % \centering
    \includegraphics[width=0.2\columnwidth]{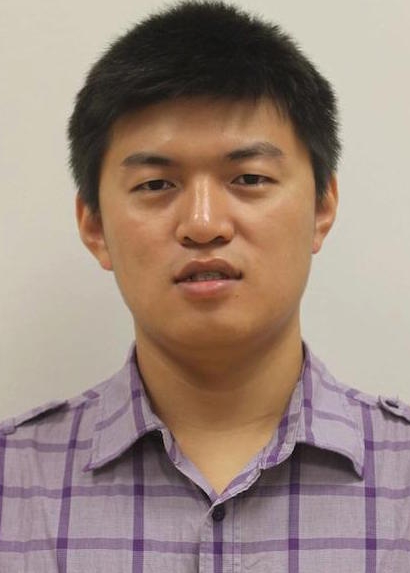}
\end{wrapfigure}
 Han Zhang received his B.S. in information science from China Agricultural University, Beijing, China, in 2009 and M.E. in communication and information systems from Beijing University of Posts and Telecommunications, Beijing, China, in 2012. He is currently a Ph.D. student of Department of Computer Science at Rutgers University, Piscataway, NJ. His current research interests include computer vision, deep learning and medical image processing.

\vspace{+5pt}
\begin{wrapfigure}{l}{0.2\columnwidth}
    % \centering
    \includegraphics[width=0.2\columnwidth]{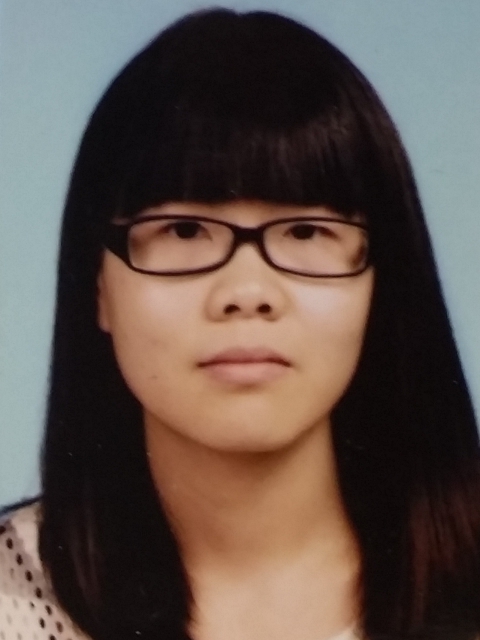}
\end{wrapfigure}
 Tao Xu received her B.E. in agricultural mechanization and automatization from China Agricultural University, Beijing, China, in 2010, and M.S. in computer science from the Institute of Computing Technology, Chinese Academy of Science, Beijing, China, in 2013. She is currently a Ph.D. student of Department of Computer Science and Engineering at Lehigh University, Bethlehem, PA. Her current research interests include deep learning, computer vision, and medical image processing.

\vspace{+5pt}
\begin{wrapfigure}{l}{0.2\columnwidth}
    % \centering
    \includegraphics[width=0.2\columnwidth]{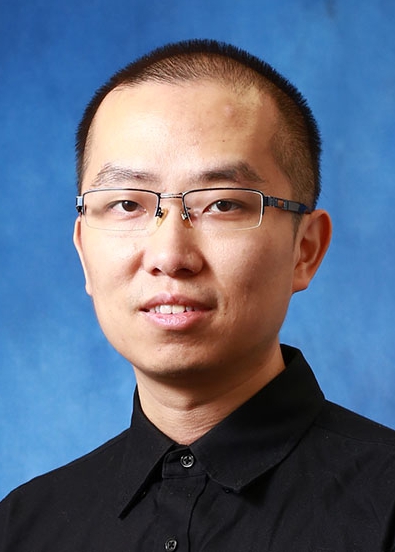}
\end{wrapfigure}
Hongsheng Li received the bachelor’s degree in automation from the East China University of Science and Technology, and the master’s and doctorate degrees in computer science from Lehigh University, Pennsylvania, in 2006, 2010, and 2012, respectively. He is currently with the Department of Electronic Engineering at the Chinese University of Hong Kong. His research interests include computer vision, medical image analysis, and machine learning.
\\

\vspace{+5pt}
\begin{wrapfigure}{l}{0.2\columnwidth}
    % \centering
    \includegraphics[width=0.2\columnwidth]{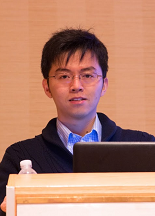}
\end{wrapfigure}
Shaoting Zhang received the BE degree from Zhejiang University in 2005, the MS degree from Shanghai Jiao Tong University in 2007, and the PhD degree in computer science from Rutgers in January 2012. His research is on the interface of medical imaging informatics, large-scale visual understanding, and machine learning. He is a senior member of the IEEE. 
\\

\vspace{+85pt}
\begin{wrapfigure}{l}{0.2\columnwidth}
    % \centering
    \includegraphics[width=0.2\columnwidth]{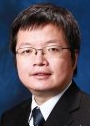}
\end{wrapfigure}
Xiaogang Wang received the BS degree from the University of Science and Technology of China in 2001, the MS degree from the Chinese University
of Hong Kong in 2003, and the PhD degree from the Computer Science and Artificial Intelligence Laboratory, Massachusetts Institute of Technology
in 2009. He is currently an associate professor in the Department of Electronic Engineering, The Chinese University of Hong Kong. His research
interests include computer vision and machine
learning. He is a member of the IEEE.

\vspace{+5pt}
\begin{wrapfigure}{l}{0.2\columnwidth}
    % \centering
    \includegraphics[width=0.2\columnwidth]{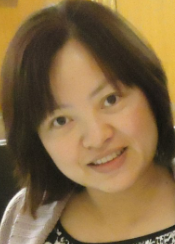}
\end{wrapfigure}
Xiaolei Huang received her doctorate degree in computer science from Rutgers University--New Brunswick, and her bachelor's degree in computer science from Tsinghua University (China). She is currently an Associate Professor in the Computer Science and Engineering Department at Lehigh University, Bethlehem, PA. Her research interests are in the areas of Computer Vision, Biomedical Image Analysis, Computer Graphics, and Machine Learning.  In these areas she has published articles in journals such as TPAMI, MedIA, TMI, TOG, and Scientific Reports. She also regularly contributes research papers to conferences such as CVPR, MICCAI, and ICCV. She serves as an Associate Editor for the Computer Vision and Image Understanding journal. She is a member of the IEEE.

\vspace{+5pt}
\begin{wrapfigure}{l}{0.2\columnwidth}
    % \centering
    \includegraphics[width=0.2\columnwidth]{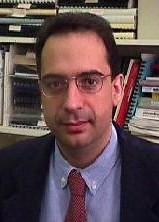}
\end{wrapfigure}
Dimitris N. Metaxas received the BE degree from the National Technical University of Athens Greece in 1986, the MS degree from the University
of Maryland in 1988, and the PhD degree from the University of Toronto in 1992. He is a professor in the Computer Science Department, Rutgers University. He is directing the Computational Biomedicine Imaging and Modeling Center
(CBIM). He has been conducting research toward the development of formal methods upon which computer vision, computer graphics, and medical imaging can advance synergistically. He is a fellow of the IEEE.

\end{document}